\documentclass[preprint,12pt]{elsarticle}

\usepackage[utf8]{inputenc} % allow utf-8 input
\usepackage[T1]{fontenc}    % use 8-bit T1 fonts

\usepackage[hyphens]{url}   % simple URL typesetting
\usepackage{booktabs}       % professional-quality tables
\usepackage{amsfonts}       % blackboard math symbols
\usepackage{nicefrac}       % compact symbols for 1/2, etc.
\usepackage{microtype}      % microtypography
\usepackage{xcolor}         % colors
\usepackage{amsmath}
\usepackage{graphicx}
\usepackage{mathtools}
\usepackage{amssymb}

\usepackage{adjustbox}
\usepackage{makecell}
\usepackage{placeins}
\usepackage{hyperref}
\usepackage[nameinlink,noabbrev]{cleveref}
\crefname{appsec}{}{}
\Crefname{appsec}{}{}
\newenvironment{kvblock}{\par\begingroup\setlength{\parindent}{0pt}\setlength{\parskip}{0pt}}{\par\endgroup}
\begin{document}

\begin{frontmatter}

\title{When Fusion Helps and When It Breaks: View-Aligned Robustness in Same-Source Financial Imaging}

\author[inst1]{Rui Ma\corref{cor1}\fnref{fn1}}
\cortext[cor1]{Corresponding author.}
\ead{marui182@mails.ucas.ac.cn}

\affiliation[inst1]{organization={Xingchong (Xi'an) Technology Co., Ltd.},
  addressline={R\&D Department},
  city={Xi'an},
  state={Shaanxi},
  country={China}}

\fntext[fn1]{ORCID: \href{https://orcid.org/0009-0002-0118-8351}{0009-0002-0118-8351}.}

\begin{abstract}
We study same-source multi-view learning and adversarial robustness for next-day direction prediction using two deterministic, window-aligned image views derived from the same time series: an OHLCV-rendered chart (\texttt{ohlcv}) and a technical-indicator matrix (\texttt{indic}).
To control label ambiguity from near-zero moves, we use an \emph{ex-post} minimum-movement threshold \texttt{min\_move} ($\tau$) based on realized $|r_{t+1}|$, defining an offline benchmark on the subset $|r_{t+1}|\ge\tau$.
Under leakage-resistant time-block splits with embargo, we compare early fusion (channel stacking) and dual-encoder late fusion with optional cross-branch consistency.
We then evaluate pixel-space $\ell_\infty$ evasion attacks (FGSM/PGD) under view-constrained and joint threat models.
We find that fusion is regime dependent: early fusion can suffer negative transfer under noisier settings, whereas late fusion is a more reliable default once labels stabilize.
Robustness degrades sharply under tiny budgets with stable view-dependent vulnerabilities; late fusion often helps under view-constrained attacks, but joint perturbations remain challenging.
\end{abstract}


\end{frontmatter}

\section{Introduction}
\label{sec:introduction}
More views do not automatically translate into better robustness.
Financial time series prediction is notoriously sensitive to label ambiguity, non-stationarity, and evaluation leakage.
In particular, next-day direction labels can be dominated by near-zero moves, where the notion of ``up'' versus ``down'' is weakly defined and easily overwhelmed by noise.
At the same time, there is growing interest in \emph{vision-style} representations of market data---candlestick charts, indicator heatmaps, and other renderings---because they enable the reuse of mature computer-vision architectures and data-augmentation pipelines.
However, when multiple views or modalities are involved, practical comparisons are often confounded by cross-source alignment issues, inconsistent timestamps, or missingness patterns, making it difficult to attribute gains to fusion design rather than data integration.

This work studies \emph{same-source multi-view learning} and adversarial robustness in a controlled setting.
We construct two deterministic, window-aligned image views from the \emph{same} underlying time series: (i) an OHLCV-rendered price/volume chart (\texttt{ohlcv}) and (ii) a technical-indicator matrix view (\texttt{indic}).
Because both views are derived from identical timestamps and a shared window definition, the setup isolates representation and fusion effects from cross-source alignment confounds.
We benchmark standard convolutional vision backbones and compare early fusion (channel stacking) against dual-encoder late fusion with a fusion head, optionally augmented with a cross-branch consistency regularizer.

A key methodological issue is that the unconditional next-day direction task can be nearly unlearnable when labels are dominated by micro-moves.
To study learnability and robustness under controlled label-ambiguity regimes, we use an \emph{ex-post} minimum-movement threshold \texttt{min\_move} ($\tau$) defined by the realized next-day return magnitude $|r_{t+1}|$.
This label-dependent filtering changes the task from unconditional direction prediction to direction prediction conditional on $|r_{t+1}|\ge\tau$; it is used solely for offline benchmarking and is not an inference-time decision rule.
To avoid leakage and keep comparisons controlled, we apply filtering once prior to splitting and then run the embargoed time-block split on the filtered index set; details are in \Cref{app:eval_details,app:eval_details}.

We further evaluate adversarial robustness under an explicit \emph{view--channel threat model}.
Using pixel-space $\ell_\infty$ attacks on the rendered inputs (FGSM/PGD)~\citep{goodfellow2014explaining,madry2017towards}, we distinguish between \emph{view-constrained} attacks that perturb exactly one view and \emph{joint} attacks that perturb both views simultaneously.
This protocol makes it possible to ask whether multi-view learning provides robustness through redundancy when only one view is corrupted, and whether such redundancy persists under coordinated perturbations.
We also conduct operating-point sensitivity checks (e.g., at $\tau=0.008$) to assess which qualitative findings are stable and which are operating-point dependent (\Cref{app:tau008}).

Our main findings, summarized in \Cref{subsec:adv_robustness} and discussed in \Cref{sec:discussion}, can be distilled as follows:
(i) fusion behavior is strongly regime dependent---early fusion can exhibit negative transfer under noisier settings, while late fusion is a more reliable default once labels stabilize;
(ii) naturally trained models degrade sharply under tiny pixel-space budgets, and robustness is markedly view dependent, revealing stable view asymmetries;
(iii) late fusion often degrades more gracefully under view-constrained attacks, but joint perturbations remain challenging; and
(iv) consistency regularization primarily improves cross-branch alignment, while its effects on clean performance and robustness are conditional and can shift with $\tau$.

\paragraph{Contributions}
We make the following contributions:
\begin{itemize}
    \item \textbf{Controlled same-source multi-view benchmark.}
    We define a window-aligned dual-view imaging protocol for SGE gold spot data (2005--2025), enabling clean attribution of fusion effects without cross-source alignment confounds.
    \item \textbf{Operating-point-aware evaluation.}
    We use \texttt{min\_move} to construct controlled label-ambiguity regimes for offline benchmarking, paired with leakage-resistant time-block splits with embargo to reduce temporal leakage (\Cref{subsec:split_embargo,app:eval_details}).
    \item \textbf{View--channel adversarial robustness analysis.}
    We evaluate FGSM/PGD under view-constrained and joint threat models, characterize sharp degradation under tiny budgets, and document stable view-dependent vulnerabilities and the conditional benefits of late fusion (\Cref{subsec:adv_robustness}).
\end{itemize}

Finally, we note the scope of this study: our baselines focus on convolutional vision backbones (Lite-CNN and ResNet18-P) and pixel-space attacks on rendered inputs.
Extending the benchmark to Transformer-based architectures and broader threat models remains an important direction for future work.

\paragraph{Organization}
After reviewing related work, we introduce the same-source multi-view setup, task, and evaluation protocol, and then specify the threat model and view-aligned robustness evaluation in \Cref{sec:threat_protocol}.
Clean baselines and fusion behavior are reported in \Cref{subsec:baseline}, followed by adversarial robustness results in \Cref{subsec:adv_robustness}.
We discuss implications and limitations in \Cref{sec:discussion} and conclude in \Cref{sec:conclusion}.
Additional implementation details and extended analyses---including the $2/255$ stress test, operating-point sensitivity at $\tau=0.008$, and full branch diagnostics---are provided in \Cref{app:stress_2over255,app:tau008,app:branch_diag_full}.

\paragraph{Code availability}
Code to reproduce the experiments will be released upon acceptance at:
\url{https://github.com/mrui166/same-source-financial-imaging-robustness}.

\section{Related Work}

\subsection{Financial Image Representations and Multi-View Fusion for Market Prediction}
\label{subsec:rw_fin_imaging_fusion}

A common line of work in market prediction converts financial time series into 2D visual representations so that convolutional models can exploit local spatial patterns and cross-variable structures.
Typical examples include candlestick-/OHLCV-style charts and chart-pattern based encodings, as well as indicator-driven matrices/heatmaps constructed from technical signals.
Beyond ad-hoc chart rendering, a more principled ``time-series-to-image'' family encodes temporal structure into images such as Gramian Angular Fields and Markov Transition Fields, enabling multi-channel image learning from a single sequence \citep{wang2015imaging}.
In financial settings, such imaging ideas have been instantiated in different ways, including technical-indicator image grids for trading \citep{sezer2019fin_barcode}, bar-/chart-window image learners \citep{sezer2019fin_barcode, jiang2023reimaging_jof}, and candlestick-pattern imaging for CNN-based recognition \citep{tsai2020candlestickcnn}.
Related work also considers incorporating cross-asset market structure via graph-enhanced predictors, e.g., a graph convolutional feature based CNN for stock trend prediction \citep{chen2021gccnn}.
Recent surveys often group multi-source fusion methods for financial market prediction by when fusion is applied---pre-model (early/data), mid-model, or post-model (late/decision)---and note that predictive performance can be sensitive to this choice \citep{fusion_fin}.
Related strands also explore learning from different representations of the \emph{same} underlying data, sometimes fusing them (e.g., feature-level fusion in hybrid CNN--RNN pipelines) \citep{kim2019featurefusion} or incorporating pattern-based representations such as candlesticks and sequence similarity \citep{liang2022candlestick_similarity}.

Within this landscape, our two-view construction follows the multi-view intuition that complementary representations can provide partially redundant evidence.
We therefore treat complementarity as an empirical question and explicitly analyze regime dependence under label-stabilization settings (via a minimum-movement threshold).
In particular, we adopt an OHLCV-rendered view that preserves price-action/volume cues and an indicator-matrix view that emphasizes trend/momentum signals; our same-source two-view design is inspired by prior multi-source image-integration approaches in financial forecasting \citep{MODII2022}, while our focus is on the same-source case to remove cross-source alignment confounds.
We focus on a \emph{same-source} setting where both views are deterministically derived from an aligned rolling window of a single market time series (SGE gold spot), which isolates the role of representation and fusion from confounding cross-source alignment issues.

A key practical takeaway from prior fusion work is that multi-view learning is not automatically beneficial: gains depend on whether the views are complementary and on careful representation alignment and fusion design \citep{Baltrusaitis2019,Rahate2022}.
More broadly, financial forecasting pipelines are also prone to subtle time leakage (e.g., through rolling normalization, window overlap, or hyperparameter search) and selection effects from repeated backtests; protocol designs such as purged CV/embargo and diagnostics such as PBO/DSR have been advocated to diagnose and quantify these risks \citep{LopezdePrado2018AFML,bailey2017pbo,bailey2014dsr}. 
Related concerns about leakage undermining benchmark validity have also been discussed in other areas (e.g., efforts to mitigate memorization-driven leakage in LLM QA benchmarks) \citep{fang2025lastingbench}, which serves as a reminder that evaluation reliability depends on stating leakage-related assumptions explicitly.
Motivated by this, our study emphasizes a controlled comparison between early fusion (channel stacking) and late fusion (dual encoders with a fusion head), and interprets gains in an \emph{evaluation-reliability--oriented} context of leakage-resistant time splits (\Cref{subsec:split_embargo}) and label-stabilization regimes (\Cref{subsec:minmove}).
Many closely related financial-imaging papers report improvements under relatively controlled evaluation, whereas we center the analysis on \emph{evaluation protocol choices} (e.g., leakage control, operating-point selection) and \emph{fusion paradigms} as first-class factors that shape reliability.

\subsection{Adversarial Robustness and Threat Models for Multi-View Inputs}
\label{subsec:rw_adv_multiview}

Adversarial robustness has been extensively studied in vision, where small $\ell_\infty$-bounded perturbations can induce systematic misclassification \citep{szegedy2014intriguing,goodfellow2014explaining}.
Canonical first-order attacks include FGSM \citep{goodfellow2014explaining} and multi-step PGD \citep{madry2017towards}, which have become standard stress tests for probing worst-case sensitivity under first-order adversaries.
Beyond generic vision benchmarks, a smaller but growing line of work considers adversarial sensitivity in \emph{financial vision} models built on chart-like representations.
For example, Chen et al.\ study adversarial robustness of a deep convolutional candlestick learner built on Gramian Angular Fields (GAF) encodings and propose perturbation sampling/adversarial training to improve stability under structured perturbations of candlestick inputs \citep{chen2020adversarial_candlestick}.
These studies are complementary to ours: they focus on candlestick-pattern classification and representation-specific attacks/defenses, whereas we evaluate next-day direction prediction under leakage-resistant time splits and explicitly compare view-constrained vs.\ joint threat models for same-source two-view fusion.
Beyond single-input settings, robustness questions also arise for multi-branch and multimodal models, where the threat model can be defined per input stream (single-source/view-constrained attacks), and where fusion architecture can modulate robustness and failure modes \citep{yang2021single_source}.
In our setting, we evaluate an inference-time, white-box, untargeted FGSM/PGD threat model instantiated as raw pixel-space $\ell_\infty$-bounded perturbations on rendered views, with optional standardized-space execution via budget conversion that preserves the underlying raw-space constraint.
We emphasize that this pixel-space threat model is a \emph{representation-level stress test} of worst-case sensitivity for the chosen rendering and fusion design, rather than a claim about a realistic financial adversary operating in the underlying market or data-generation process.

Time-series settings introduce additional subtleties: perturbations can be defined either in the raw temporal domain or on downstream representations (e.g., rendered images), and the choice determines what ``small'' means and which invariances are being tested.
Prior work has instantiated adversarial attacks directly on time-series classifiers and shown that standard gradient-based methods can transfer to temporal signals \citep{fawaz2019advtsc}.
In financial pipelines, where inputs often undergo deterministic preprocessing (windowing, indicator computation, normalization), robustness evaluation therefore benefits from threat models that explicitly account for \emph{where} perturbations are applied and \emph{how} multi-view inputs are aligned.

Our work aligns with this perspective by treating pixel-space $\ell_\infty$ perturbations on rendered views as a representation-level stress test, and by formalizing the multi-view attack surface through view--channel mapping.
This enables (i) view-constrained perturbations that target exactly one view in two-view models and (ii) joint perturbations that attack both views, while maintaining a view-aligned comparison to single-view baselines.
Such protocols help disentangle robustness contributions from (a) the chosen representation, (b) early vs.\ late fusion, and (c) auxiliary alignment objectives (e.g., cross-view consistency), while reducing confounds from data-splitting artifacts or cross-source misalignment.

\section{Problem Setup and Notation}
\label{sec:problem_setup}

\subsection{Same-Source Data (SGE Gold Spot)}
\begin{kvblock}
\textbf{Source.} Shanghai Gold Exchange (SGE), spot gold; data retrieved via Tushare \citep{tushare}. \\
\textbf{Time range.} 2005-09-26--2025-08-28 (all dates reported in UTC). \\
\textbf{Granularity.} Trading-day level observations (non-trading days excluded; trading days follow the SGE calendar). \\
\textbf{Fields.} Price and trading activity information (OHLC and trading volume). \\
\textbf{Preprocessing.} We remove invalid records, sort the data chronologically, and construct rolling lookback windows. Both image views used in our experiments are generated from the same window and aligned to the same window end date.
\end{kvblock}

\subsection{Multi-View Construction}
Motivated by the multi-view formulation in MODII~\citep{MODII2022}, we construct two complementary 2D image views from the same market-data window: (i) an OHLCV-based price/volume view (\texttt{ohlcv}) and (ii) a technical-indicator view (\texttt{indic}).
Both views are generated from the same reference day and aligned to the same 15-day lookback window, enabling controlled comparisons across single-view and two-view settings.
Our construction follows a MODII-style pipeline~\citep{MODII2022} and is conceptually related to prior time-series-to-image transformations for financial prediction~\citep{sezer2019fin_barcode}.
Rendering, scaling, and ordering details are provided in \ref{app:view_render}.

\paragraph{OHLCV view (\texttt{ohlcv})}
We render OHLC bars with a volume subpanel to capture recent level, range, and trading activity in a compact visual representation.

\paragraph{Indicator view (\texttt{indic})}
We compute a fixed set of $N=15$ standard indicators over the same lookback window and convert the resulting indicator-by-time matrix into a grayscale image; the indicator set and parameters are fixed \emph{a priori} (\ref{app:view_render}).

\begin{figure}[!htbp]
  \centering
    \includegraphics[width=0.18\linewidth]{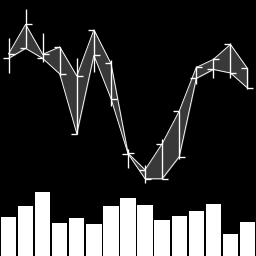}
    \hspace{0.12\linewidth}
    \includegraphics[width=0.18\linewidth]{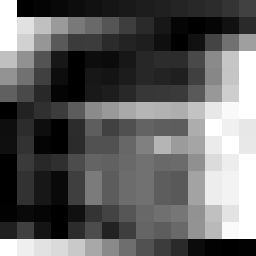}
    \caption{Examples of \texttt{ohlcv} and \texttt{indic} constructed from a fixed lookback window of $L_{\mathrm{ohlcv}}=L_{\mathrm{indic}}=15$ trading days.}
  \label{fig:indic_examples}
\end{figure}

\subsection{View-to-Channel Mapping}\label{sec:view-to-channel-mapping}
For two-view models (\texttt{both}), we concatenate the two grayscale images along the channel dimension,
\begin{equation}
\mathbf{x}=[v_{\texttt{ohlcv}};v_{\texttt{indic}}]\in\mathbb{R}^{2\times256\times256},
\end{equation}
with \texttt{ch0}$\!=\!\texttt{ohlcv}$ and \texttt{ch1}$\!=\!\texttt{indic}$.
For single-view baselines, the provided view is always placed in \texttt{ch0}, yielding $\mathbf{x}\in\mathbb{R}^{1\times256\times256}$.
This convention enables channel-constrained perturbations to target a specific view in the robustness protocol (\Cref{sec:threat_protocol}).

\subsection{Task and Evaluation Metrics}

\paragraph{Prediction target}
For each trading day $t$, we form an input sample by extracting a lookback window of the past $L$ trading days,
$\{t-L+1,\dots,t\}$, and predict the next-day directional movement at $(t+1)$. In all experiments, we set $L=15$.
We define the next-day simple return as
\begin{equation}
\label{eq:next_day_return}
r_{t+1}=\frac{close_{t+1}-close_t}{close_t}, \qquad
y_t=\mathbb{I}\!\left(r_{t+1}>0\right),
\end{equation}
where $close_t$ denotes the closing price on day $t$ and $\mathbb{I}(\cdot)$ is the indicator function.
Thus, the task is a binary classification problem with labels $y_t\in\{0,1\}$.
When using the ex-post movement threshold \texttt{min\_move} ($\tau$), we report performance on the conditional subset $|r_{t+1}|\ge\tau$; see \ref{app:eval_details} for the filtering/split protocol.

\paragraph{Matthews Correlation Coefficient (MCC)}
We use MCC as the primary metric because it summarizes the full confusion matrix and remains informative under class imbalance \citep{matthews1975comparison}.
MCC can be interpreted as a correlation between predicted and true labels, taking values in $[-1,1]$.
\begin{equation}
\mathrm{MCC}=\frac{TP\cdot TN-FP\cdot FN}
{\sqrt{(TP+FP)(TP+FN)(TN+FP)(TN+FN)}}.
\end{equation}
We report mean$\pm$std over multiple fixed random seeds (\ref{app:eval_details}).

\section{Threat Model and Evaluation Protocol}
\label{sec:threat_protocol}

We evaluate test-time, white-box evasion robustness on the rendered image inputs using standard pixel-space $\ell_\infty$ perturbations and first-order attacks~\citep{szegedy2014intriguing,goodfellow2014explaining,madry2017towards}.
Our goal is a view-aligned stress test: (i) perturb exactly one view for two-view models (view-constrained) or (ii) perturb both views jointly, while keeping comparisons to single-view baselines well defined via the view--channel mapping (\Cref{sec:view-to-channel-mapping}).

\subsection{Threat Model}
\label{sec:threat_model}

\paragraph{$\ell_\infty$-bounded perturbations in raw pixel space}
Let $\mathbf{x}_{\mathrm{raw}}\in[0,1]^{C\times256\times256}$ denote the raw input (with $C=1$ for single-view and $C=2$ for two-view).
An adversary produces $\mathbf{x}'_{\mathrm{raw}}=\mathbf{x}_{\mathrm{raw}}+\boldsymbol{\delta}$ subject to~\citep{szegedy2014intriguing,goodfellow2014explaining,madry2017towards}
\begin{equation}
\label{eq:linf}
\|\boldsymbol{\delta}\|_\infty \le \epsilon_{\mathrm{adv}},
\qquad \|\boldsymbol{\delta}\|_\infty \triangleq \max_i |\delta_i|.
\end{equation}

\paragraph{Budgets}
We evaluate a small-budget grid
\begin{equation}
\label{eq:eps_grid_main}
E=\left\{0,\ \frac{0.25}{255},\ \frac{0.5}{255},\ \frac{0.75}{255},\ \frac{1}{255}\right\},
\end{equation}
where $\epsilon_{\mathrm{adv}}=0$ corresponds to clean evaluation.

\paragraph{Attacks}
We use FGSM and multi-step PGD as untargeted cross-entropy attacks~\citep{goodfellow2014explaining,madry2017towards}.
All attack hyperparameters (steps, step size, projection, and clipping) are reported in \ref{app:attack_params}.

\paragraph{Standardization and valid-range handling}
The threat model is always defined in raw $[0,1]$ space.
When models consume standardized inputs, we execute attacks in the standardized space using budget conversion that preserves the underlying raw-space constraint, and clip to the corresponding valid range (\ref{app:input_param}).

\subsection{Threat Scenarios and View-Aligned Protocol}
\label{sec:protocol_main}

Using the view--channel mapping (\Cref{sec:view-to-channel-mapping}), we define two threat scenarios.

\paragraph{View-constrained perturbations (single-view corruption)}
For two-view inputs, the adversary perturbs exactly one view/channel~\citep{yang2021single_source}:
\begin{equation}
\label{eq:view_constrained_main}
\boldsymbol{\delta}=(\delta_{\texttt{ch0}},\mathbf{0}) \ \text{(attack \texttt{ohlcv})}
\quad \text{or} \quad
\boldsymbol{\delta}=(\mathbf{0},\delta_{\texttt{ch1}}) \ \text{(attack \texttt{indic})}.
\end{equation}

\paragraph{Joint perturbations (coordinated corruption)}
For two-view inputs, the adversary perturbs both channels:
\begin{equation}
\label{eq:joint_main}
\boldsymbol{\delta}=(\delta_{\texttt{ch0}},\delta_{\texttt{ch1}}).
\end{equation}

\paragraph{View-aligned comparison to single-view baselines}
Single-view models always place the provided view in \texttt{ch0}; therefore, we align comparisons by the attacked \emph{view}:
\texttt{ohlcv} (single-view \texttt{ohlcv} attacked on \texttt{ch0}) vs.\ (\texttt{both} attacked on \texttt{ch0}),
and \texttt{indic} (single-view \texttt{indic} attacked on \texttt{ch0}) vs.\ (\texttt{both} attacked on \texttt{ch1}).
Joint perturbations apply only to two-view models.

\subsection{Evaluation and Summary Metrics}
\label{sec:eval_main}

We report robustness curves $\mathrm{MCC}(\epsilon_{\mathrm{adv}})$ over $\epsilon_{\mathrm{adv}}\in E$ (Eq.~\ref{eq:eps_grid_main}), with mean$\pm$std over fixed training seeds; details on probability-to-label conversion and reporting are in \ref{app:eval_details}.
To summarize robustness across budgets, we report the worst-case degradation over nonzero budgets:
\begin{equation}
\label{eq:worst_main}
\Delta(\epsilon_{\mathrm{adv}})=\mathrm{MCC}(\epsilon_{\mathrm{adv}})-\mathrm{MCC}(0),\qquad
\Delta_{\mathrm{worst}}=\min_{\epsilon_{\mathrm{adv}}\in E,\ \epsilon_{\mathrm{adv}}>0}\Delta(\epsilon_{\mathrm{adv}}).
\end{equation}

\paragraph{Late-fusion diagnostics (analysis only)}
For late-fusion two-view models, we additionally report branch-level diagnostics on the clean test set to aid interpretation; see \ref{app:branch_diag_full}.

\section{A Stable, Leakage-Resistant Training Baseline}
We consider a set of learning and non-learning baselines to evaluate our image representations under a unified, leakage-resistant evaluation protocol.
All models perform binary next-day direction prediction and output logits/probabilities.
Inputs are either a single view (\texttt{indic} or \texttt{ohlcv}) or a two-view input (\texttt{both}); for early-fusion models, the two resized views are stacked channel-wise (\Cref{sec:view-to-channel-mapping}).
Unless otherwise stated, trainable baselines share the same optimization and checkpoint-selection protocol (\ref{app:train_proto}).

\subsection{Model Architectures}

\paragraph{Compact CNN baseline (\texttt{Lite-CNN})}
We use a lightweight convolutional network adapted from the MODII framework~\citep{MODII2022}.
In \texttt{both} mode, we use early fusion by channel-wise stacking of the two views; in single-view mode, the model consumes one channel.

\paragraph{Pretrained ResNet-18 baseline (\texttt{ResNet18-P})}
We employ a ResNet-18 backbone initialized with ImageNet pretrained weights~\citep{he2016deep,deng2009imagenet}.
We adapt only the first convolution to match the required number of input channels; the rest of the backbone is unchanged.
Two-view \texttt{both} uses early fusion (stacked channels) as above.

\paragraph{Dual-view consistency variants (\texttt{*-cons})}
For two-view inputs, we additionally evaluate late-fusion dual-branch variants for both backbones, denoted \texttt{Lite-CNN-Cons} and \texttt{ResNet18-P-Cons}.
These models produce view-specific logits and a fused-logit output; training uses the objective in \Cref{subsec:consistency-regularization}.
We report $\lambda\in\{0,0.5,1\}$ (with $\lambda=0$ reducing to fused-logit cross-entropy only) and fix $T=2.0$ unless stated otherwise.

\paragraph{Logistic Regression baseline (\texttt{LogReg})}
We apply logistic regression on flattened image features to measure how far linear decision boundaries on the constructed representations can go.

\paragraph{Majority-class baseline (\texttt{Majority})}
We include a non-learning baseline that always predicts the most frequent class in the \emph{training} split.

For reproducibility, implementation details of backbones and fusion heads are provided in \ref{app:model_arch}, including a concise implementation-level summary in \Cref{tab:impl_backbone_fusion}.

\subsection{Cross-view Consistency Regularization}
\label{subsec:consistency-regularization}

\paragraph{Motivation}
In the two-view setting (\texttt{both}), each sample provides two aligned views of the same market state and shares the same label.
We encourage the two view-specific predictors to produce coherent predictions via an explicit consistency regularizer.

\paragraph{Dual-branch setup}
Let the two branches produce logits $z_a$ and $z_b$ from the two views, and let a fusion head produce fused logits $z_f$ from concatenated branch features.
The fused logits $z_f$ define the supervised objective; $z_a$ and $z_b$ define the consistency term.

\paragraph{Temperature-scaled symmetric KL}
We define temperature-scaled predictive distributions~\citep{hinton2015distilling,zhang2018deepmutual}
\begin{equation}
p_a=\mathrm{softmax}(z_a/T),\qquad p_b=\mathrm{softmax}(z_b/T),
\end{equation}
and use the symmetric KL divergence
\begin{equation}
\label{eq:cons}
\mathcal{L}_{\mathrm{cons}}(z_a,z_b)
=\frac{T^2}{2}\Big[\mathrm{KL}(p_a\,\|\,p_b)+\mathrm{KL}(p_b\,\|\,p_a)\Big].
\end{equation}

\paragraph{Training objective}
We optimize
\begin{equation}
\mathcal{L}=\mathcal{L}_{\mathrm{CE}}(z_f,y)+\lambda\,\mathcal{L}_{\mathrm{cons}}(z_a,z_b),
\end{equation}
where $\lambda\ge 0$ controls the regularization strength.
Branch-level diagnostics used for interpretation are reported in \ref{app:branch_diag_full}.

\subsection{Stabilizing Labels with a Minimum-Movement Threshold}
\label{subsec:minmove}

Next-day direction labels can be ambiguous when the realized price change is extremely small.
To study how label ambiguity affects fusion and robustness, we optionally define an \emph{ex-post} evaluation subset using a minimum-movement threshold $\tau$:
for each sample ending at day $t$, we compute $r_{t+1}$ (Eq.~\ref{eq:next_day_return}) and keep the sample only if $|r_{t+1}|\ge\tau$.
We evaluate $\tau\in\{0,\,0.002,\,0.004,\,0.006,\,0.008,\,0.010\}$ (denoted \texttt{min\_move}).

\paragraph{Protocol note}
Because inclusion depends on the realized $|r_{t+1}|$, \texttt{min\_move} defines a conditional offline benchmark rather than an inference-time decision rule.
To keep comparisons controlled across split variants, filtering is applied before splitting so that all variants operate on the same subset definition (\ref{app:eval_details}).

\subsection{Leakage-Resistant Time Split with Embargo}
\label{subsec:split_embargo}

Because inputs are built from overlapping sliding windows, random splitting can introduce look-ahead leakage.
We therefore adopt a leakage-resistant time-block split with embargo for all models~\citep{LopezdePrado2018AFML}.
Trading days are grouped into consecutive blocks of size $B$ (\texttt{block\_size}); earlier blocks form train, then validation, and the latest blocks form test (default 70/15/15 by blocks).

To mitigate boundary leakage across split boundaries, we apply an embargo of at least $D_{\mathrm{emb}}$ trading days (\texttt{embargo\_days})~\citep{LopezdePrado2018AFML,Racine2000HVBlock}.
Implementation details and the sample-size impact under our default configuration (\texttt{block\_size}=30, \texttt{embargo\_days}=20) are reported in \ref{app:eval_details} (\Cref{tab:minmove_embargo_effect}).

\subsection{Implementation Details}
\label{subsec:impl_details}

Training and selection details (optimizer, early stopping, and reproducibility controls) are provided in \ref{app:train_proto}.
Input standardization and leakage-avoiding normalization are detailed in \ref{app:input_param}, and adversarial-attack hyperparameters are summarized in \ref{app:attack_params}.

\section{Results}
\subsection{Clean Baselines}
\label{subsec:baseline}

\begin{figure}[!htbp]
  \centering
  \includegraphics[width=0.98\linewidth]{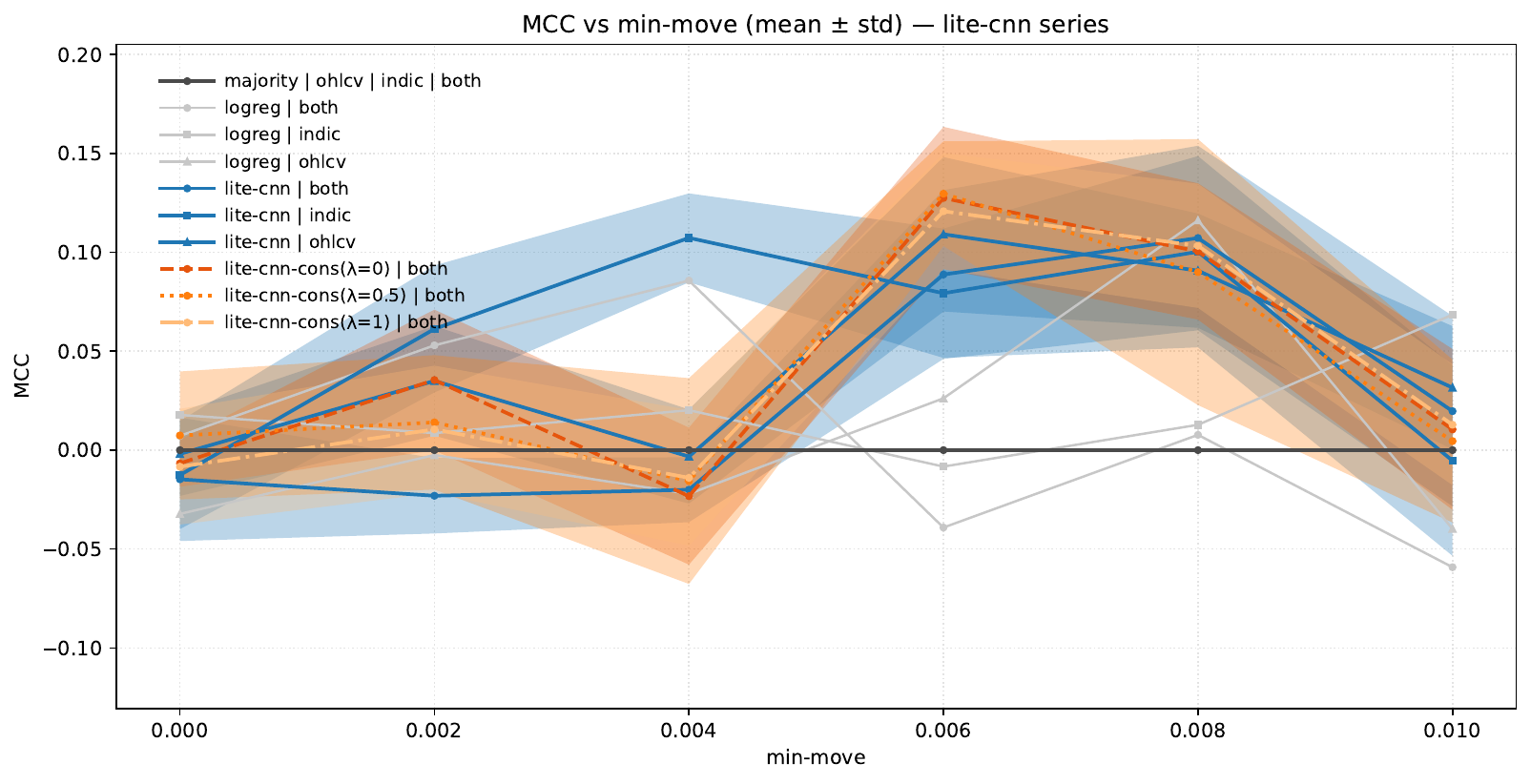}\par\vspace{1mm}
  \includegraphics[width=0.98\linewidth]{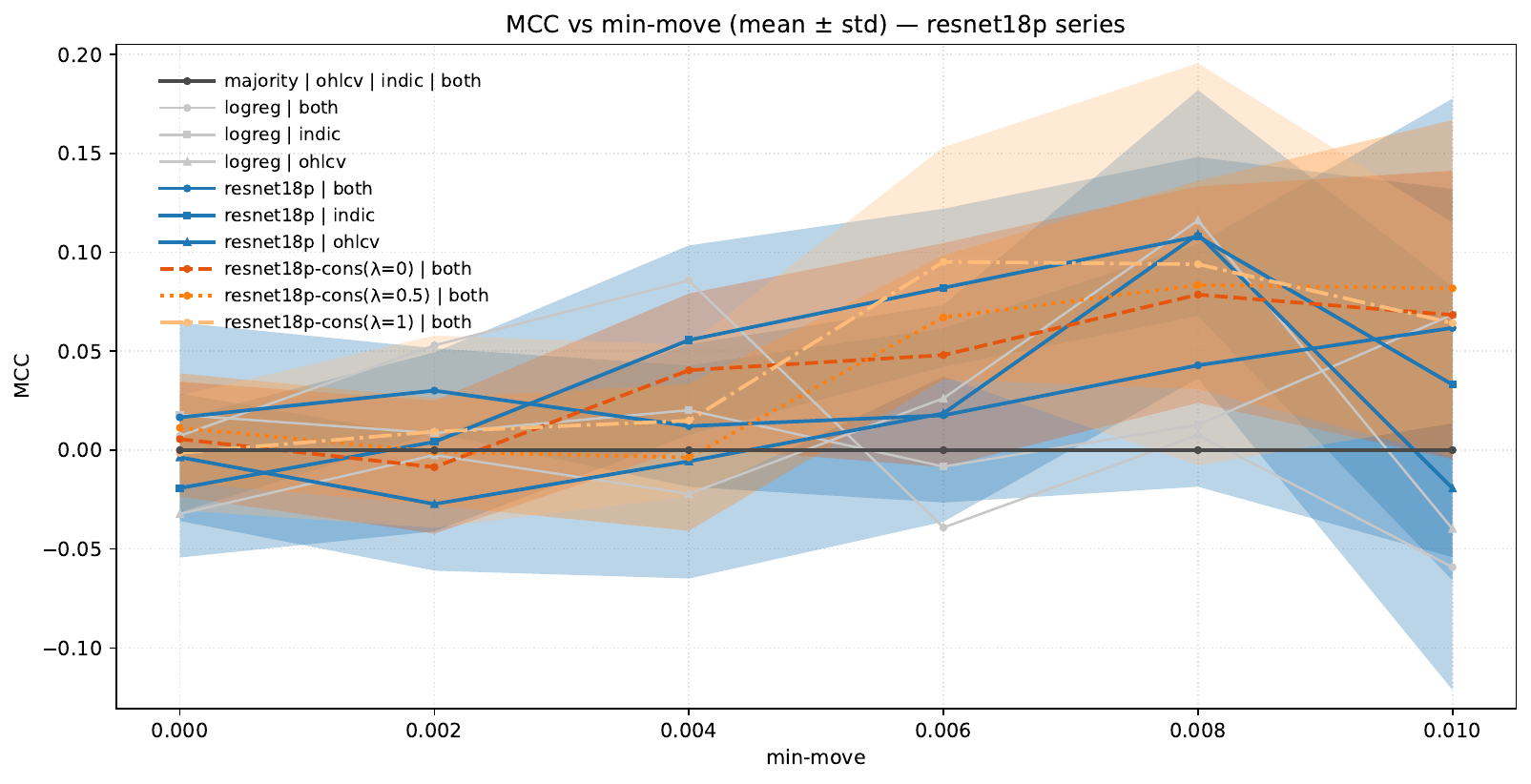}
  \caption{\textbf{Baseline MCC vs.\ minimum-movement threshold $\tau$ (mean$\pm$std).}
  Top: Lite-CNN family. Bottom: ResNet18-P family.
  Deep models are averaged over $n{=}8$ random seeds; Majority and LogReg are run once ($n{=}1$).
  }
  \label{fig:baseline_minmove}
\end{figure}

\paragraph{Setup and evaluation metric}
We compare Majority, LogReg, and two CNN backbones (Lite-CNN and ResNet18-P) under three input modes: \texttt{ohlcv}, \texttt{indic}, and dual-view \texttt{both}.
For \texttt{both}, vanilla CNN baselines use early fusion (channel-wise stacking), whereas \texttt{*-cons} variants use a dual-branch late-fusion head and optionally add cross-view consistency regularization weighted by $\lambda$.
We use MCC as the primary decision metric since it accounts for all TP/TN/FP/FN entries and penalizes degenerate single-direction predictors.

\paragraph{Minimum-movement filtering determines the effective difficulty}
As shown in \Cref{fig:baseline_minmove}, performance is near chance without filtering or under weak filtering ($\tau\in\{0,0.002,0.004\}$), but improves sharply once $\tau$ reaches the mid range (notably $\tau\in\{0.006,0.008\}$), consistent with reduced label ambiguity on the ex-post filtered subsets.
Aggregating across all deep configurations (Lite-CNN/ResNet18-P and \texttt{*-cons}, all input modes), the mean MCC rises from $-0.002$ at $\tau=0$ to $0.082$ at $\tau=0.006$ and peaks at $0.092$ at $\tau=0.008$, before dropping to $0.030$ at $\tau=0.010$.
The drop at $\tau=0.010$ aligns with the reduced effective sample size and increased variance (see the sample-count/embargo summary in \Cref{tab:minmove_embargo_effect}, in \Cref{app:eval_details}).
We stress that $\tau$ defines offline evaluation regimes ex post, not an inference-time filtering rule; the filtering/splitting protocol is specified in \Cref{app:eval_details}.

\paragraph{Fusion behavior once labels stabilize}
Once $\tau\ge 0.006$, dual-view information becomes exploitable and \texttt{both} turns consistently positive for Lite-CNN family; however, under weak filtering (e.g., $\tau=0.004$) early fusion can exhibit negative transfer relative to the stronger single-view option (\Cref{tab:baseline_mcc_key}).
Across the two-view setting, late fusion is the dominant architectural gain over early fusion (representative numbers in \Cref{tab:baseline_mcc_key}), while consistency regularization yields modest, $\tau$-dependent and non-monotonic effects.

\paragraph{Backbone differences}
ResNet18-P exhibits a noisier regime: dual-view early fusion is generally weak relative to single-view modes at the same $\tau$, and late fusion partially recovers performance but with higher variance, especially under stricter filtering (see \Cref{fig:baseline_minmove} and representative values in \Cref{tab:baseline_mcc_key}).

\paragraph{LogReg as a diagnostic}
LogReg can be competitive at certain thresholds but is non-monotonic across $\tau$ and input modes; we treat it primarily as a diagnostic indicator of partial linear separability after label stabilization.

\paragraph{Takeaways}
Across backbones, (i) label stabilization via \texttt{min\_move} is a prerequisite for learning in our setting; (ii) in the dual-view setting, late fusion is generally more reliable than early fusion on clean evaluation once labels stabilize; and (iii) consistency regularization can provide modest gains in some mid-noise regimes but is not reliably monotonic and requires careful tuning jointly with $\tau$.
For full numerical commentary (without changing any conclusions or values), see \Cref{app:clean_baselines_full}.

\begin{table}[!htbp]
  \caption{\textbf{Representative MCC (mean$\pm$std) at key $\tau$ values.}
  Deep models are averaged over $n{=}8$ seeds. Full curves are in \Cref{fig:baseline_minmove}.}
  \centering
  \small
  \setlength{\tabcolsep}{6pt}
  \begin{adjustbox}{max width=\textwidth,center}
\begin{tabular}{lccc}
    \toprule
    \textbf{Model (input)} & $\boldsymbol{\tau=0.004}$ & $\boldsymbol{\tau=0.006}$ & $\boldsymbol{\tau=0.008}$ \\
    \midrule
    Lite-CNN (\texttt{both}, early fusion) & $-0.020004\pm0.016396$ & $0.088767\pm0.042662$ & $\mathbf{0.107206\pm0.046632}$ \\
    Lite-CNN (\texttt{indic})             & $\mathbf{0.107292\pm0.022477}$  & $0.079303\pm0.032678$ & $0.100279\pm0.048322$ \\
    Lite-CNN (\texttt{ohlcv})             & $-0.003236\pm0.023789$ & $0.109095\pm0.039023$ & $0.090707\pm0.028893$ \\
    Lite-CNN-Cons (\texttt{both}, $\lambda=0$)   & $-0.023262\pm0.034602$ & $0.127459\pm0.035946$ & $0.100430\pm0.034439$ \\
    Lite-CNN-Cons (\texttt{both}, $\lambda=0.5$) & $-0.015634\pm0.052001$ & $\mathbf{0.129607\pm0.026449}$ & $0.089973\pm0.067209$ \\
    Lite-CNN-Cons (\texttt{both}, $\lambda=1$)   & $-0.014013\pm0.034309$ & $0.120755\pm0.029468$ & $0.103269\pm0.031387$ \\
    \midrule
    ResNet18-P (\texttt{both}, early fusion) & $0.012168\pm0.030674$ & $0.017593\pm0.044072$ & $0.042873\pm0.061370$ \\
    ResNet18-P (\texttt{indic})             & $0.055638\pm0.047824$ & $0.082039\pm0.039988$ & $0.108023\pm0.040145$ \\
    ResNet18-P (\texttt{ohlcv})             & $-0.005656\pm0.059240$ & $0.018580\pm0.054761$ & $0.109101\pm0.072830$ \\
    ResNet18-P-Cons (\texttt{both}, $\lambda=0.5$) & $-0.003653\pm0.037078$ & $0.067112\pm0.031388$ & $0.083423\pm0.052584$ \\
    \bottomrule
  \end{tabular}
\end{adjustbox}
  \label{tab:baseline_mcc_key}
\end{table}

\paragraph{Choice of $\tau$ for robustness evaluation}
All subsequent adversarial evaluations use $\tau=0.006$ as the default operating point.
At smaller thresholds ($\tau\le 0.004$), most configurations remain close to chance in MCC, making robustness comparisons less informative.
At larger thresholds (e.g., $\tau=0.010$), the reduced effective sample size can amplify variance across random seeds.
As a sensitivity check, we repeat the key robustness evaluations at $\tau=0.008$ and observe the same qualitative conclusions (see \Cref{app:tau008}).
Thus, $\tau=0.006$ offers a practical balance between label stabilization and sufficient data for reliable robustness measurement.

\subsection{Adversarial Robustness}
\label{subsec:adv_robustness}

\begin{figure}[!htbp]
  \centering
  \includegraphics[width=0.32\linewidth]{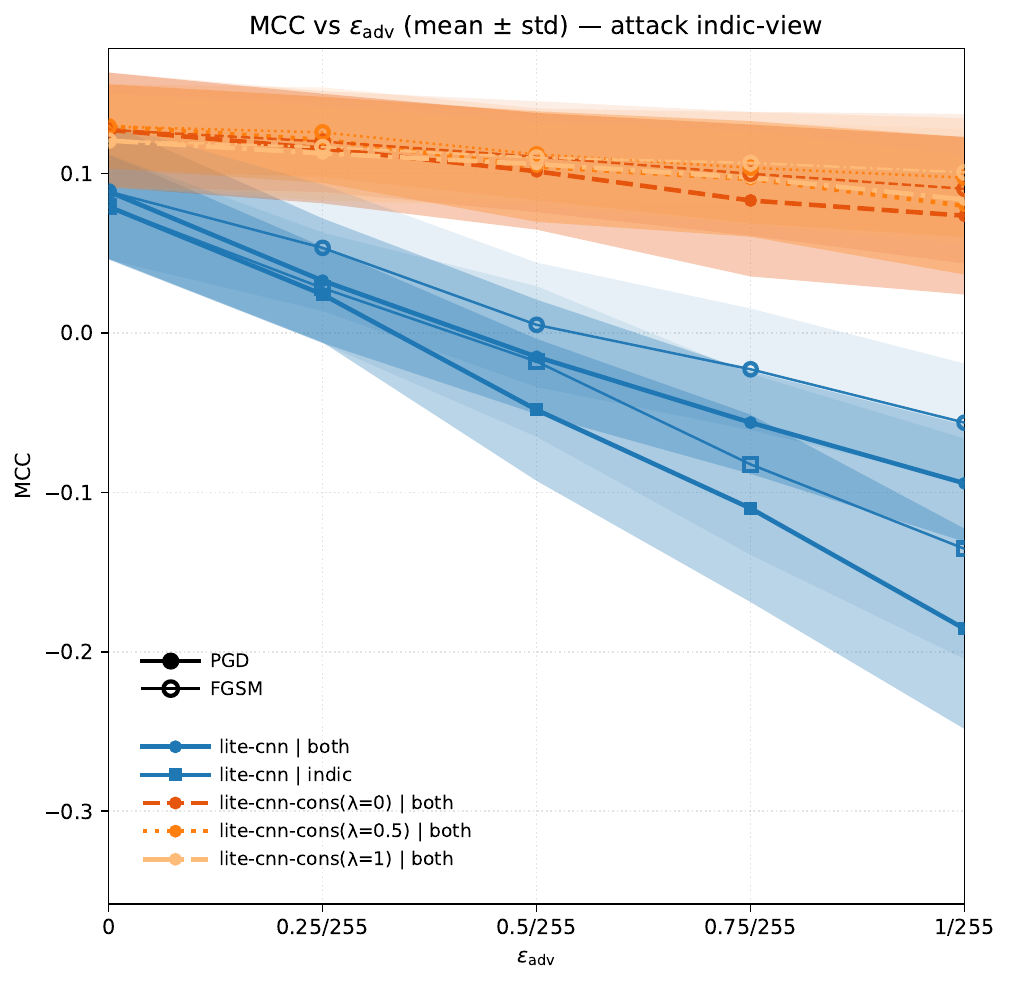}\hfill
  \includegraphics[width=0.32\linewidth]{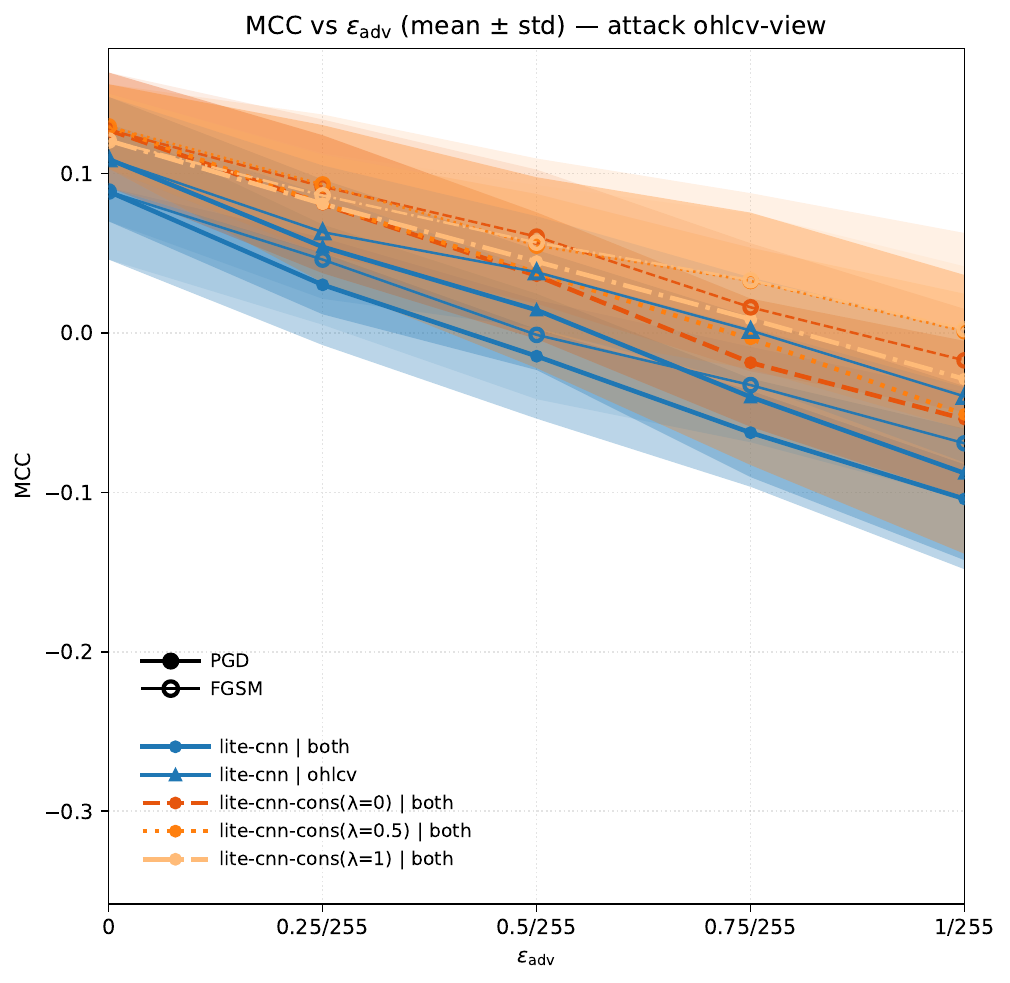}\hfill
  \includegraphics[width=0.32\linewidth]{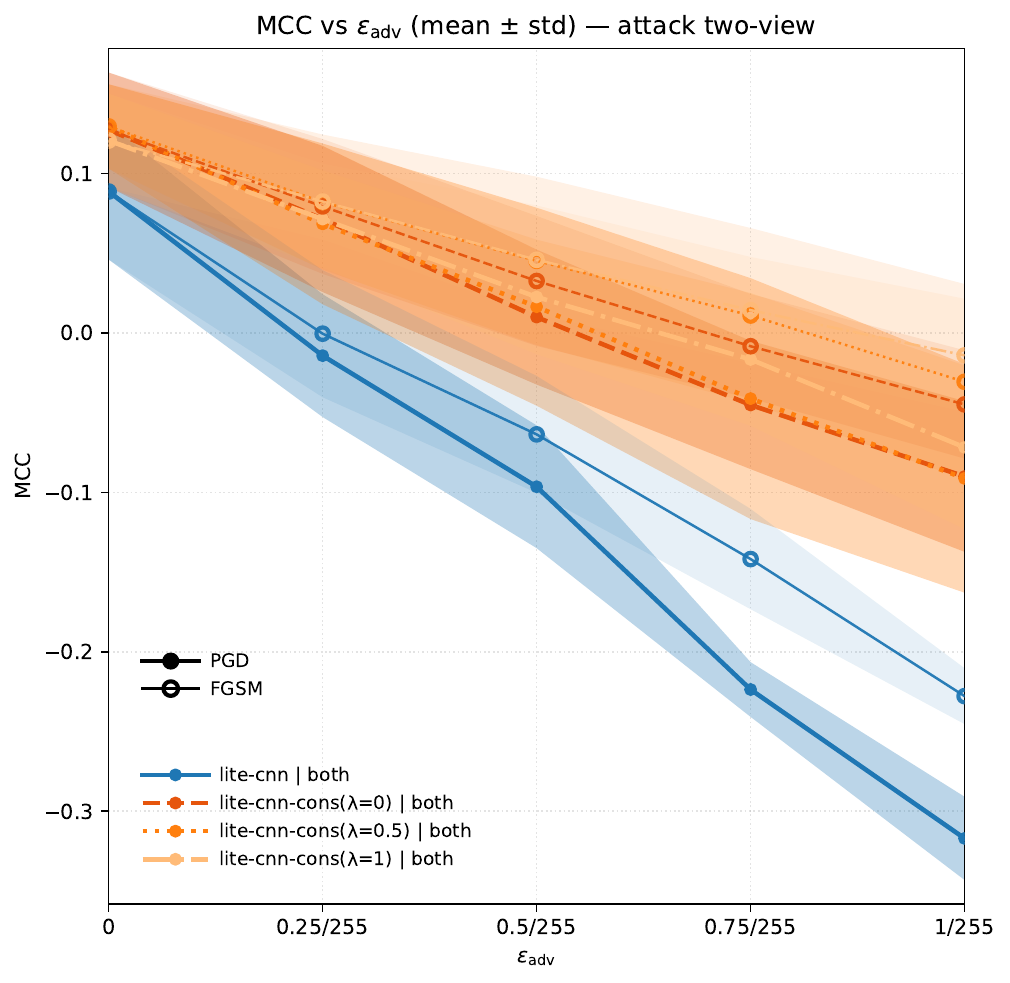}
  \caption{\textbf{Adversarial robustness curves for Lite-CNN family at $\tau=0.006$ (mean$\pm$std).}
  We report $\mathrm{MCC}(\epsilon_{\mathrm{adv}})$ under FGSM and PGD for (left) attacking the \texttt{indic} view,
  (middle) attacking the \texttt{ohlcv} view, and (right) joint perturbations on both views.
  Late fusion (\texttt{*-cons}) degrades more gracefully than early fusion, especially under view-constrained attacks.}
  \label{fig:adv_curves_litecnn}
\end{figure}

\begin{figure}[!htbp]
  \centering
  \includegraphics[width=0.32\linewidth]{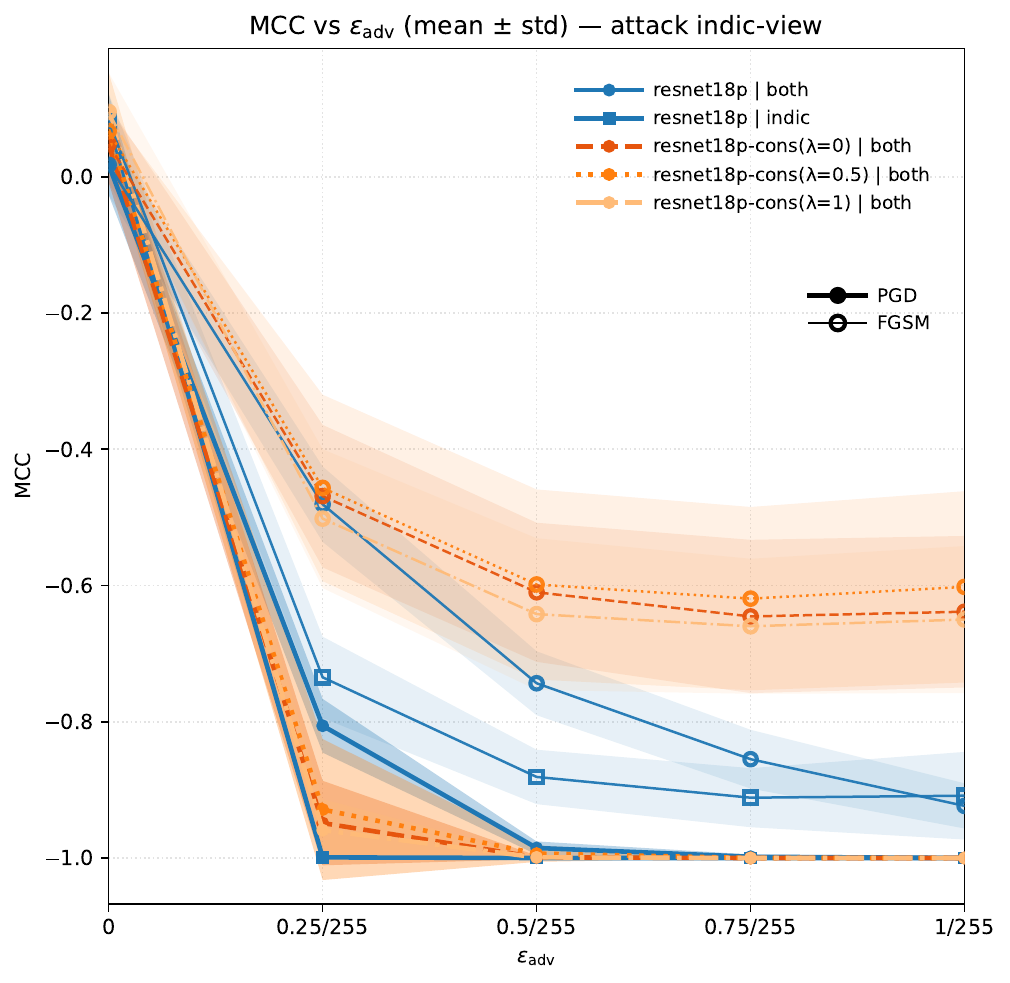}\hfill
  \includegraphics[width=0.32\linewidth]{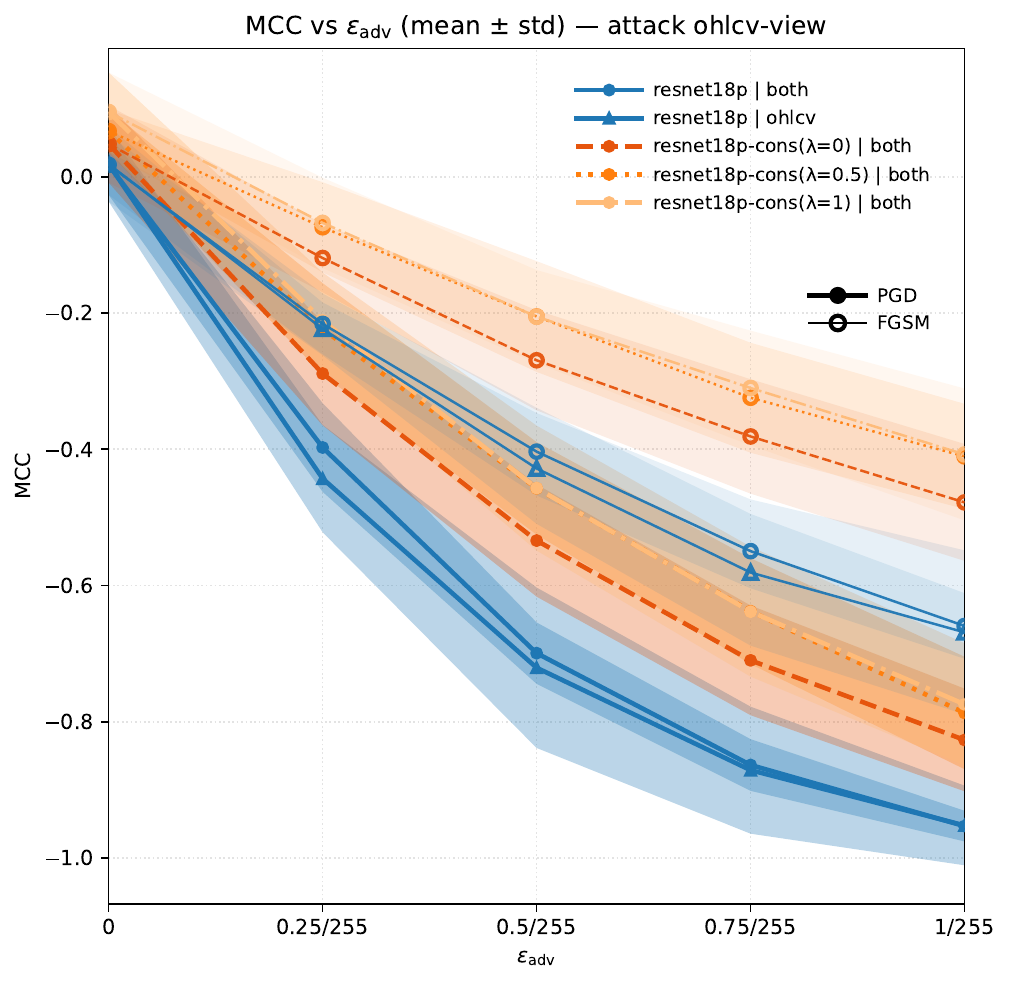}\hfill
  \includegraphics[width=0.32\linewidth]{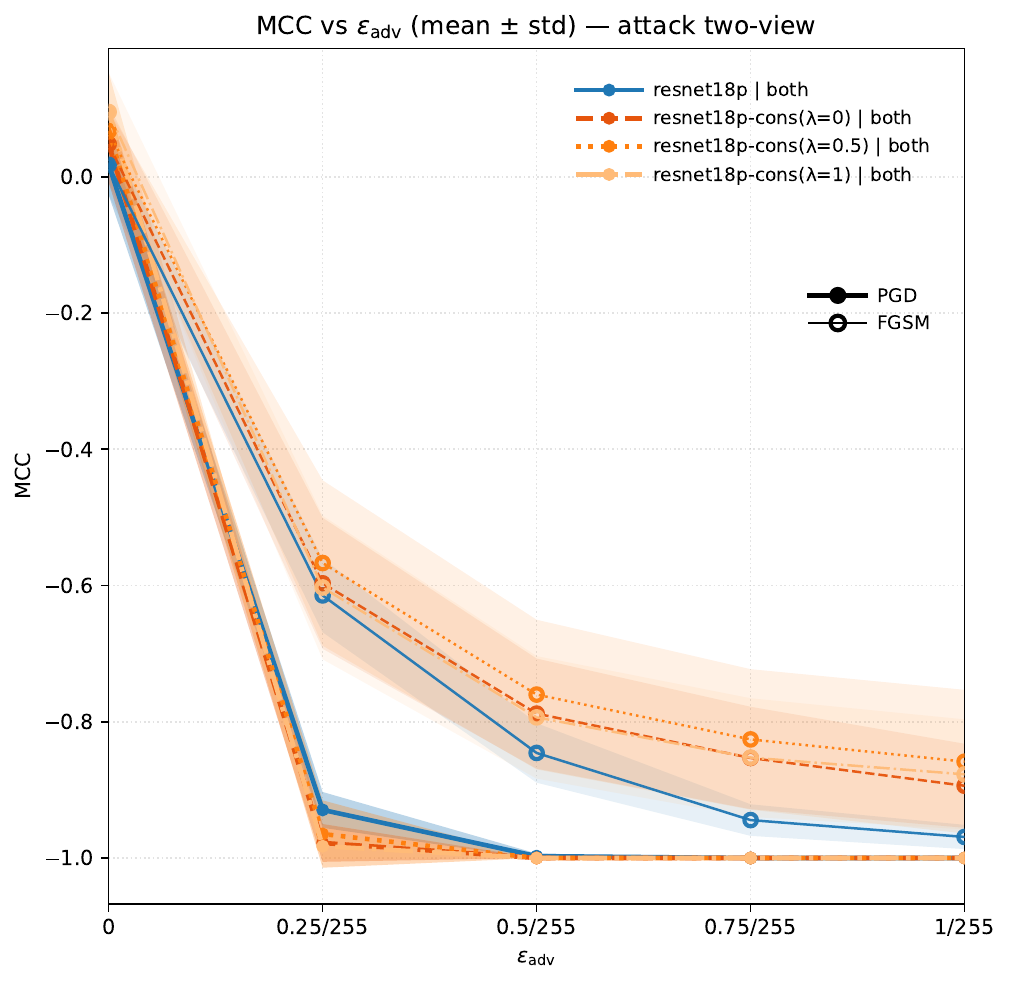}
  \caption{\textbf{Adversarial robustness curves for ResNet18-P family at $\tau=0.006$ (mean$\pm$std).}
  We report $\mathrm{MCC}(\epsilon_{\mathrm{adv}})$ under FGSM and PGD for (left) attacking the \texttt{indic} view,
  (middle) attacking the \texttt{ohlcv} view, and (right) joint perturbations on both views.
  The family shows pronounced view sensitivity (indicator attacks are particularly destructive), and late fusion only partially mitigates this.}
  \label{fig:adv_curves_resnet18p}
\end{figure}

\paragraph{Protocol and summary metrics}
We evaluate test-time evasion attacks under the $\ell_\infty$ threat model in Eq.~\eqref{eq:linf} using the main budget grid $E$ in Eq.~\eqref{eq:eps_grid_main} (up to $\epsilon_{\mathrm{adv}}=1/255$).
Unless stated otherwise, all robustness results in this subsection use the default operating point $\tau=0.006$ (min\_move) selected in \Cref{subsec:baseline}.
A sensitivity check at $\tau=0.008$ is reported in \Cref{app:tau008}, and the $2/255$ stress test in \Cref{app:stress_2over255}.
We report robustness curves $\mathrm{MCC}(\epsilon_{\mathrm{adv}})$ (mean$\pm$std over random seeds), and summarize degradation by
$\Delta_{\mathrm{worst}}$ as defined in Eq.~\eqref{eq:worst_main}.
We focus on PGD as the primary evaluation attack; FGSM is included as a weaker, single-step reference.
\Cref{fig:adv_curves_litecnn,fig:adv_curves_resnet18p} visualize these robustness curves for the Lite-CNN and ResNet18-P families.

\paragraph{Summary of findings}
At the default operating point $\tau=0.006$, we observe (i) rapid degradation under tiny pixel-space budgets,
(ii) pronounced view dependence that is backbone- and fusion-dependent, and (iii) a robustness advantage of late fusion under \emph{view-constrained} attacks,
while robustness under \emph{joint} (two-view) attacks remains challenging.
The sensitivity study at $\tau=0.008$ in \Cref{app:tau008} confirms the structural trends in (i) and (iii), but indicates that finer-grained \emph{hyperparameter} conclusions
(e.g., the best $\lambda$ within late fusion) can shift with the operating point.
In contrast, the \emph{direction} of view asymmetry is primarily model- and fusion-dependent: ResNet18-P is consistently most vulnerable to \texttt{indic}-only perturbations,
whereas Lite-CNN-Cons in late fusion is consistently more vulnerable to \texttt{ohlcv}-only perturbations under view-constrained attacks (\Cref{tab:adv_summary_pgd,tab:adv_summary_pgd_tau008}).


\paragraph{Overall vulnerability under tiny $\ell_\infty$ budgets}
Across architectures, naturally trained models can degrade substantially under pixel-space perturbations.
Even at $\epsilon_{\mathrm{adv}}=0.25/255$ (a quarter of an 8-bit intensity level), PGD often causes a pronounced MCC drop, and performance can become near-zero or negative as the budget increases toward $1/255$.
This confirms that pixel-space $\ell_\infty$ robustness is not obtained ``for free'' in our financial-image setting.

\paragraph{View sensitivity and view-aligned evaluation}
Using the view--channel mapping in \Cref{sec:view-to-channel-mapping}, we compare (i) single-view models attacked on their sole channel and
(ii) two-view models attacked on the corresponding view channel (\Cref{sec:protocol_main}).
The results reveal strong view dependence, but its direction is backbone- and fusion-dependent.
For ResNet18-P, perturbing the indicator view is consistently the most destructive at both $\tau=0.006$ and $\tau=0.008$, and very small budgets can already drive MCC close to the lower bound, suggesting near-saturated failure rather than a gradual attenuation of predictive signal.
For Lite-CNN variants, view asymmetry is milder and depends on the fusion pathway; notably, for Lite-CNN-Cons with late fusion, \texttt{ohlcv}-only perturbations are consistently the most destructive single-view attacks at both $\tau$ values (\Cref{tab:adv_summary_pgd,tab:adv_summary_pgd_tau008}).

\paragraph{Late fusion improves robustness under view-constrained attacks}
A consistent trend in the two-view setting is that dual-branch late fusion (\texttt{*-cons}) degrades more gracefully than early fusion (channel stacking) under view-constrained attacks.
When only one view is perturbed, the unperturbed branch can still provide usable evidence, which mitigates worst-case collapse and yields a flatter robustness curve over $E$.
This advantage is most pronounced in the Lite-CNN family under single-view attacks (and persists at $\tau=0.008$; \Cref{app:tau008}).
For ResNet18-P, late fusion improves robustness against OHLCV-only perturbations but does not prevent saturated failures under indicator-only attacks.

\paragraph{Joint perturbations remain challenging}
When the adversary perturbs both channels jointly, performance deteriorates more rapidly than in view-constrained settings, even for late fusion.
This indicates that redundancy across views provides meaningful protection against single-view corruption, but does not fully defend against worst-case joint perturbations.

\paragraph{Consistency regularization: alignment, not a universal robustness knob}
Within late-fusion models, the consistency weight $\lambda$ primarily enforces cross-branch logit alignment, while its impact on fused-head robustness is secondary and can be non-monotonic across operating points.
We therefore treat $\lambda$ as a tunable refinement on top of the late-fusion design rather than a universally optimal setting; detailed branch diagnostics supporting this interpretation are provided in \Cref{app:branch_diag_full}.

\begin{table}[!htbp]
  \caption{\textbf{PGD robustness summary at $\tau=0.006$ (min\_move).}
  We report MCC at two representative budgets ($0.25/255$ and $1/255$) and the worst-case drop $\Delta_{\mathrm{worst}}$ over nonzero budgets $\{0.25,0.5,0.75,1\}/255$ (Eq.~\eqref{eq:worst_main}).
  As an additional stress test, results at $\epsilon_{\mathrm{adv}}=2/255$ are reported in \Cref{app:stress_2over255}.}
  \label{tab:adv_summary_pgd}
  \centering
  \small
  \setlength{\tabcolsep}{5pt}
  \begin{adjustbox}{max width=\textwidth,center}
\begin{tabular}{lllrrrr}
    \toprule
    \textbf{Model} & \textbf{Input} & \textbf{Attack (PGD)} &
    \textbf{MCC@0} & \textbf{MCC@0.25/255} & \textbf{MCC@1/255} & $\boldsymbol{\Delta_{\mathrm{worst}}}$ \\
    \midrule
    Lite-CNN & \texttt{ohlcv} & on \texttt{ohlcv} & 0.109 & 0.054 & -0.088 & -0.197 \\
    Lite-CNN & \texttt{indic} & on \texttt{indic} & 0.079 & 0.024 & -0.186 & -0.265 \\
    \midrule
    Lite-CNN & \texttt{both} (early) & on \texttt{ohlcv} (ch0) & 0.089 & 0.030 & -0.104 & -0.193 \\
    Lite-CNN & \texttt{both} (early) & on \texttt{indic} (ch1) & 0.089 & 0.033 & -0.094 & -0.183 \\
    Lite-CNN & \texttt{both} (early) & on both & 0.089 & -0.014 & -0.317 & -0.406 \\
    \midrule
    Lite-CNN-Cons ($\lambda{=}1$) & \texttt{both} (late) & on \texttt{ohlcv} (ch0) & 0.121 & 0.081 & -0.029 & -0.150 \\
    Lite-CNN-Cons ($\lambda{=}1$) & \texttt{both} (late) & on \texttt{indic} (ch1) & 0.121 & 0.113 & 0.084 & -0.037 \\
    Lite-CNN-Cons ($\lambda{=}1$) & \texttt{both} (late) & on both & 0.121 & 0.071 & -0.072 & -0.192 \\
    \midrule
    ResNet18-P & \texttt{ohlcv} & on \texttt{ohlcv} & 0.019 & -0.443 & -0.952 & -0.971 \\
    ResNet18-P & \texttt{indic} & on \texttt{indic} & 0.082 & -0.999 & -1.000 & -1.082 \\
    \midrule
    ResNet18-P-Cons ($\lambda{=}1$) & \texttt{both} (late) & on \texttt{ohlcv} (ch0) & 0.095 & -0.211 & -0.774 & -0.870 \\
    ResNet18-P-Cons ($\lambda{=}1$) & \texttt{both} (late) & on \texttt{indic} (ch1) & 0.095 & -0.959 & -1.000 & -1.095 \\
    ResNet18-P-Cons ($\lambda{=}1$) & \texttt{both} (late) & on both & 0.095 & -0.983 & -1.000 & -1.095 \\
    \bottomrule
  \end{tabular}
\end{adjustbox}
\end{table}

\paragraph{Takeaways}
Across all settings, pixel-space $\ell_\infty$ perturbations can severely degrade next-day direction prediction even at extremely small budgets.
Robustness is strongly view-dependent and architecture-specific (ResNet18-P is particularly fragile to \texttt{indic}-only attacks).
In the two-view regime, dual-branch late fusion generally improves robustness under \emph{view-constrained} attacks, supporting the intuition that an unperturbed branch can provide a partial evidence path.
However, \emph{joint} perturbations remain challenging and can still induce large performance drops.
Finally, consistency regularization mainly controls cross-branch alignment; its effect on robustness is secondary and can depend on the operating point (\Cref{app:tau008}).

\paragraph{PGD stress test at $\epsilon_{\mathrm{adv}}=2/255$ (not in the main grid)}
We extend the perturbation budget to $\epsilon_{\mathrm{adv}}=2/255$ for late-fusion two-view Lite-CNN-Cons models and find that they remain \emph{non-trivially robust} under the view-constrained \texttt{indic}-only attack (ch1), while \texttt{ohlcv}-only and joint attacks are substantially more destructive; full results at both $\tau=0.006$ and $\tau=0.008$ are reported in \Cref{app:stress_2over255}.

\section{Discussion}
\label{sec:discussion}

\paragraph{Operating point, noise regime, and fusion}
All findings should be interpreted relative to the operating point induced by \texttt{min\_move} ($\tau$) in \Cref{subsec:baseline}.
Increasing $\tau$ reduces direction-label ambiguity by filtering near-zero moves but also reduces effective sample size and can increase variance.
In noisy regimes (weak filtering), early fusion by channel stacking can underperform the stronger single-view baseline, consistent with negative transfer; once labels stabilize (around $\tau\ge 0.006$ in our setup), using both views becomes more reliably beneficial and dual-branch late fusion is a robust default.
Sensitivity at $\tau=0.008$ in \Cref{app:tau008} preserves the qualitative patterns, while showing that some within-family hyperparameter preferences can shift with the operating point.

\paragraph{Robustness under tiny budgets: redundancy helps only under view-constrained attacks}
Across backbones and operating points, naturally trained models can degrade sharply under small pixel-space $\ell_\infty$ budgets (\Cref{subsec:adv_robustness,tab:adv_summary_pgd,tab:adv_summary_pgd_tau008}).
Late fusion often degrades more gracefully under \emph{view-constrained} attacks, consistent with preserving an unperturbed evidence path when only one view is perturbed.
This mitigation is limited: view-specific failure modes can still saturate quickly, and \emph{joint} perturbations remain challenging for both early and late fusion, including in the $2/255$ stress test (\Cref{app:stress_2over255}).

\paragraph{Stable view asymmetry as a diagnostic}
Vulnerability is strongly view dependent.
ResNet18-P is consistently most fragile to indicator-view perturbations, while Lite-CNN-Cons under late fusion is more sensitive to \texttt{ohlcv}-only perturbations under view-constrained attacks.
This suggests that robustness depends on view rendering and backbone inductive bias, and the direction of view asymmetry provides a useful diagnostic for interpreting view-aligned robustness results and identifying view-specific failure modes.

\paragraph{Consistency regularization and limitations}
Branch diagnostics in \Cref{app:branch_diag_full} show that increasing $\lambda$ improves cross-branch agreement, but robustness effects are conditional: the clean-optimal $\lambda$ is backbone-dependent at $\tau=0.006$ (\Cref{subsec:baseline}), and robustness rankings among $\lambda\in\{0,0.5,1\}$ can change with $\tau$ (\Cref{app:tau008}).
Our study evaluates pixel-space attacks on rendered images (a representation-level stress test), uses a label-dependent subset definition for offline benchmarking (\Cref{app:eval_details}), and benchmarks only convolutional backbones; broader model families and threat models remain valuable extensions.

\section{Conclusion}
\label{sec:conclusion}

We studied same-source multi-view learning and adversarial robustness for next-day direction prediction using two deterministic, window-aligned image views (\texttt{ohlcv} and \texttt{indic}), enabling explicit view--channel threat modeling.
Under leakage-resistant time-block splits with embargo, learnability and robustness depend strongly on the label-ambiguity regime induced by near-zero moves.
We therefore use an \emph{ex-post} minimum-movement filter (\texttt{min\_move}) to define conditional evaluation subsets, revealing a non-monotonic data--noise trade-off: moderate $\tau$ improves learnability on the remaining subset, while overly strict filtering reduces sample size and increases variance.
Importantly, \texttt{min\_move} is a label-dependent subset definition for offline benchmarking and is not an inference-time decision rule (\Cref{app:eval_details}).

In stabilized regimes (default $\tau=0.006$), early fusion can suffer negative transfer under noisier settings, whereas dual-branch late fusion is a more reliable default and provides the dominant clean-performance gains (\Cref{subsec:baseline}).
From a robustness perspective, naturally trained models can be highly vulnerable to pixel-space $\ell_\infty$ perturbations even at tiny budgets (\Cref{subsec:adv_robustness}).
Late fusion often degrades more gracefully under view-constrained attacks, but joint perturbations remain challenging, indicating that cross-view redundancy alone does not guarantee worst-case robustness (\Cref{app:stress_2over255}).
Sensitivity at $\tau=0.008$ preserves the qualitative conclusions while highlighting operating-point dependence in within-family hyperparameters (\Cref{app:tau008}).
Overall, reliable robustness assessment in same-source financial imaging benefits from leakage-resistant temporal evaluation, explicit reporting of operating points and label-dependent subset definitions, and threat models that distinguish view-constrained from joint multi-view attacks.

%%%%%%%%%%%%%%%%%%%%%%%%%%%%%%%%%%%%%%%%%%%%%%%%%%%%%%%%%%%%
\FloatBarrier

\section*{Declaration of competing interests}
The author declares no competing interests.

\bibliographystyle{elsarticle-num}
\bibliography{references}

\newpage\appendix
\crefalias{section}{appsec}

\section{Implementation Details for Input Parameterization}
\label{app:input_param}

\Cref{tab:input_param} summarizes the input parameterization and normalization settings.

\paragraph{Raw input}
Each view is stored as an 8-bit grayscale image and loaded with \texttt{ToTensor()}, yielding a raw input
$\mathbf{x}_{\mathrm{raw}}\in[0,1]^{C\times256\times256}$,
where $C=1$ for single-view and $C=2$ for two-view models.

\paragraph{Optional per-channel standardization}
Optionally, we apply per-channel standardization
\begin{equation}
\label{eq:app_standardize}
\mathbf{x}_{\mathrm{norm}}=(\mathbf{x}_{\mathrm{raw}}-\boldsymbol{\mu})\oslash\boldsymbol{\sigma},
\end{equation}
where $(\boldsymbol{\mu},\boldsymbol{\sigma})$ are estimated by randomly sampling
$\min(n_{\text{norm}},|{\rm train}|)$ images from the training split (default $n_{\text{norm}}=512$) without replacement,
computing per-channel statistics over $N\times H\times W$, and clamping $\sigma$ to be at least $10^{-6}$.
When \texttt{use\_norm} is enabled, the model consumes $\mathbf{x}_{\mathrm{norm}}$; otherwise, it consumes $\mathbf{x}_{\mathrm{raw}}$.

\paragraph{Valid-range clipping}
Clipping is performed in the space where the attack is executed:
raw inputs are clipped to $[0,1]$; standardized inputs are clipped to
$\big[(0-\boldsymbol{\mu})\oslash\boldsymbol{\sigma},\ (1-\boldsymbol{\mu})\oslash\boldsymbol{\sigma}\big]$.

\paragraph{Threat model defined in raw space}
The threat model is always defined on $\mathbf{x}_{\mathrm{raw}}$ with an $\ell_\infty$ constraint
$\|\boldsymbol{\delta}\|_\infty \le \epsilon_{\mathrm{adv}}$,
where $\|\boldsymbol{\delta}\|_\infty = \max_i|\delta_i|$.
When the model consumes standardized inputs, we enforce the same underlying raw-space budget by converting
\begin{equation}
\label{eq:app_eps_convert}
\epsilon_{\mathrm{norm}}=\epsilon_{\mathrm{adv}} \oslash \boldsymbol{\sigma},
\qquad
\alpha_{\mathrm{norm}}=\alpha \oslash \boldsymbol{\sigma},
\end{equation}
and then performing clipping/projection in the standardized space. This preserves the raw-space bound after mapping back.

\begin{table}[!htbp]
\caption{Input parameterization and normalization details used in our implementation.}
\centering
\label{tab:input_param}
\small
\begin{adjustbox}{max width=\textwidth,center}
\begin{tabular}{@{}l p{0.72\linewidth}@{}}
\toprule
\textbf{Item} & \textbf{Setting} \\
\midrule
Image storage & 8-bit grayscale per view \\
Load / scaling & \texttt{ToTensor()} $\Rightarrow \mathbf{x}_{\mathrm{raw}}\in[0,1]^{C\times256\times256}$ \\
Normalization (\texttt{use\_norm}) & $\mathbf{x}_{\mathrm{norm}}=(\mathbf{x}_{\mathrm{raw}}-\mu)\oslash\sigma$ \\
$\mu,\sigma$ estimation &
Randomly sample $\min(n_{\text{norm}},|{\rm train}|)$ training images without replacement (default $n_{\text{norm}}{=}512$);
compute per-channel mean/std over $N{\times}H{\times}W$; clamp $\sigma \leftarrow \max(\sigma,10^{-6})$ \\
Clipping (raw) & $[0,1]$ \\
Clipping (norm) & $\big[(0-\mu)\oslash\sigma,(1-\mu)\oslash\sigma\big]$ \\
Threat model space & raw space; when attacking norm, use Eq.~\ref{eq:app_eps_convert} \\
\bottomrule
\end{tabular}
\end{adjustbox}
\end{table}

\section{View Rendering and Multi-View Construction Details}
\label{app:view_render}

This appendix details the view construction summarized in \Cref{sec:problem_setup}.
We follow the MODII-style design for multi-view financial image construction~\citep{MODII2022} and adopt standard time-series-to-image principles commonly used in financial prediction pipelines~\citep{sezer2019fin_barcode}.

\paragraph{OHLCV rendering (\texttt{ohlcv})}
Each sample is rendered on a black canvas at a fixed resolution of $256\times256$ pixels.
The upper region (75\% of the height) visualizes OHLC as candlestick-style glyphs (a high--low vertical stroke with open/close ticks).
We shade the open--close range in gray to reflect the interval-valued view of price movements~\citep{Han2012Interval}.
The lower region (25\% of the height) encodes trading volume as a bar chart.
Volumes are normalized by a rolling percentile scaler, i.e., divided by the trailing 95th percentile computed over a fixed reference window of length $L_{\mathrm{vol}}$, and then clipped to $[0,1]$.

\paragraph{Indicator-matrix rendering (\texttt{indic})}
We compute a fixed set of $N=15$ technical indicators over a lookback window of $L_{\mathrm{indic}}=15$ trading days.
Concretely, we use RSI(14), WILLR(14), WMA(10), SMA(15), EMA(12), DEMA(14), TEMA(18), CCI(20), CMO(14), MACD(hist; 8,17,9), PPO(8,17; EMA), ROC(10), MFI(14), ADX(14), and SAR(0.02, 0.2); all choices and hyperparameters are fixed \emph{a priori}.
For each sample, we stack the most recent $L_{\mathrm{indic}}$ values into an $N\times L_{\mathrm{indic}}$ matrix (rows: indicators; columns: time) and apply per-row min--max scaling to $[0,255]$ within the current window.
To keep the spatial layout consistent across splits, we compute a single indicator row order on the training split via average-linkage hierarchical clustering on the absolute Spearman correlation (distance $1-|\rho|$) with optimal leaf ordering (falling back to a spectral ordering if unavailable), and reuse it for validation/test. :contentReference[oaicite:0]{index=0} :contentReference[oaicite:1]{index=1}
Finally, the $15\times 15$ matrix is upsampled with nearest-neighbor interpolation to $256\times256$ (to match the image encoder input size) and rendered as a grayscale image.

\section{Model Architecture and Fusion Details}
\label{app:model_arch}

\Cref{tab:impl_backbone_fusion} provides an implementation-level summary of all backbones and fusion heads used in the main experiments; the remainder of this appendix specifies layer-wise and fusion-head details for reproducibility.

\begin{table}[!htbp]
\caption{\textbf{Implementation-level summary of model backbones and fusion heads.}
Here $C$ is the number of input channels (1 for single-view; 2 for early-fusion two-view). For late fusion, each branch consumes one channel and the fused head operates on concatenated features.}
\label{tab:impl_backbone_fusion}
\centering
\small
\setlength{\tabcolsep}{4pt}
\renewcommand{\arraystretch}{1.08}
\begin{adjustbox}{max width=\textwidth,center}
\begin{tabular}{@{}p{0.20\linewidth}p{0.37\linewidth}p{0.37\linewidth}@{}}
\toprule
\textbf{Component} & \textbf{Lite-CNN (PaperCNN)} & \textbf{ResNet18-P} \\
\midrule
Single-view encoder ($C{=}1$) &
$3$ conv blocks + pools; MLP head (256, 128) with ReLU; Dropout after first FC &
ResNet-18 trunk (512-d) + Dropout + Linear$(512\!\rightarrow\!2)$ \\

Early fusion two-view ($C{=}2$) &
Same network with $C{=}2$ at input (channel stacking) &
Same network with $C{=}2$ at input (adapted first conv; scale-corrected) \\

Late fusion encoders (\texttt{*-Cons}) &
Two PaperCNN trunks ($C{=}1$ each), output $d{=}128$ &
Two ResNet-18 trunks ($C{=}1$ each), output $d{=}512$ \\

Fused head (\texttt{*-Cons}) &
Dropout + Linear$(256\!\rightarrow\!2)$ &
Dropout + Linear$(1024\!\rightarrow\!2)$ \\

Branch heads (for consistency) &
Linear$(128\!\rightarrow\!2)$ for each branch (no extra dropout in head) &
Dropout + Linear$(512\!\rightarrow\!2)$ for each branch \\
\bottomrule
\end{tabular}
\end{adjustbox}
\end{table}

\paragraph{Notation}
All models output binary logits. For single-view inputs, $C{=}1$; for two-view inputs, $C{=}2$ where the two views are stacked channel-wise. In late-fusion models, the input is split as $\mathbf{x}_a=\mathbf{x}[:,0{:}1]$ and $\mathbf{x}_b=\mathbf{x}[:,1{:}2]$.

\paragraph{Lite-CNN (PaperCNN)}
Our Lite-CNN baseline is a compact CNN with three convolution blocks and two pooling operations, followed by a 2-layer MLP feature head and a final linear classifier. Concretely:
(1) $\mathrm{Conv}_{3\times3}(C\!\rightarrow\!16)$ + ReLU;
(2) $\mathrm{Conv}_{3\times3}(16\!\rightarrow\!16)$ + ReLU + $\mathrm{MaxPool}(2,2)$;
(3) $\mathrm{Conv}_{3\times3}(16\!\rightarrow\!32)$ + ReLU + $\mathrm{MaxPool}(3,3)$;
then flatten and an MLP/classifier head
$\mathrm{Linear}(\cdot\!\rightarrow\!256)\!\rightarrow\!\mathrm{ReLU}\!\rightarrow\!\mathrm{Dropout}\!\rightarrow\!\mathrm{Linear}(256\!\rightarrow\!128)\!\rightarrow\!\mathrm{ReLU}\!\rightarrow\!\mathrm{Linear}(128\!\rightarrow\!2)$.
The flatten dimension is inferred automatically from the input resolution (256$\times$256).

\paragraph{ResNet18-P}
We use a ResNet-18 backbone initialized from ImageNet weights. The first convolution layer is adapted to accept $C\in\{1,2\}$ channels using the pretrained 3-channel weights with a scale correction:
for $C{=}1$, we use the channel-wise mean; for $C{=}2$ (more generally, $C{<}3$), we copy the first $C$ kernels and multiply by $(3/C)$ to stabilize activation magnitudes; for $C{>}3$, we repeat kernels along the channel dimension and apply the same $(3/C)$ scale factor.
We split the network into a \emph{trunk} (up to global average pooling, producing a 512-d feature) and a \emph{head} (Dropout + Linear$(512\!\rightarrow\!2)$).

\paragraph{Late fusion with optional consistency (\texttt{*-Cons})}
For two-view late fusion, we use two independent encoders (\emph{branches}) that each process one view:
$\mathbf{f}_a=g_a(\mathbf{x}_a)$ and $\mathbf{f}_b=g_b(\mathbf{x}_b)$.
The \emph{fused head} predicts from concatenated features
$\mathbf{f}=[\mathbf{f}_a;\mathbf{f}_b]$ via Dropout + Linear$(2d\!\rightarrow\!2)$ (where $d{=}128$ for PaperCNN trunks and $d{=}512$ for ResNet trunks).

For consistency-regularized models, each branch also outputs auxiliary logits used only for the consistency term, while the main task loss is computed on the fused logits.
In the ResNet18-P variant, the auxiliary branch heads are Dropout + Linear$(d\!\rightarrow\!2)$ (i.e., the per-branch head modules).
In the PaperCNN variant, the auxiliary branch heads are Linear$(d\!\rightarrow\!2)$ without an extra dropout layer (dropout is applied inside the trunk before producing $\mathbf{f}_a,\mathbf{f}_b$).

\section{Additional Evaluation Details}
\label{app:eval_details}

\paragraph{From probabilities to labels}
Unless stated otherwise, we obtain binary predictions by thresholding the predicted positive-class probability at $0.5$ and compute MCC from the resulting confusion matrix.

\paragraph{Reporting}
We report mean$\pm$std over $n{=}8$ runs with different random seeds.
Unless otherwise stated, the seed set is fixed to $\{1,2,3,4,5,6,7,8\}$ for all deep models to enable controlled comparisons across ablations.
For non-learning (Majority) and deterministic baselines (LogReg), we run once ($n{=}1$) unless explicitly noted.

\paragraph{Filtering protocol for \texttt{min\_move}}
When using \texttt{min\_move} ($\tau$), we restrict evaluation to the conditional subset defined by $|r_{t+1}|\ge\tau$.
Because inclusion depends on the realized next-day return, \texttt{min\_move} is an \emph{offline} benchmark definition rather than an inference-time decision rule.
To keep comparisons controlled across split variants and avoid inconsistencies near split boundaries, we apply the filter once on the full timeline prior to splitting, and then run the time-block split with embargo on the filtered index set (\Cref{subsec:split_embargo}, \ref{app:eval_details}).
The resulting sample-size impact across $\tau$ values is summarized in \Cref{tab:minmove_embargo_effect}.

\paragraph{Time-Block Split and Embargo}
This paragraph complements \Cref{subsec:split_embargo} with implementation details of the time-block split with embargo~\citep{LopezdePrado2018AFML,Racine2000HVBlock}.

Let $B$ denote the block size (\texttt{block\_size}) and $D_{\mathrm{emb}}$ the embargo length in trading days (\texttt{embargo\_days}).
We drop $K=\lceil D_{\mathrm{emb}}/B\rceil$ adjacent blocks around split boundaries to ensure a temporal gap of at least $D_{\mathrm{emb}}$ days (possibly larger due to block granularity)~\citep{LopezdePrado2018AFML,Racine2000HVBlock}.
\Cref{tab:minmove_embargo_effect} reports, for each \texttt{min\_move} setting, the aligned sample counts before splitting and the final counts retained after applying the embargoed time-block split, together with the resulting embargo percentage.
All reported results use the same split protocol across models and settings.

\begin{table}[!htbp]
\centering
\caption{Impact of \texttt{min\_move} filtering and embargo under the time-block split protocol (\texttt{block\_size}=30, \texttt{embargo\_days}=20). \emph{Aligned} denotes samples after filtering and view alignment, and \emph{Used} denotes samples retained after applying the embargo split.}
\label{tab:minmove_embargo_effect}
\begin{adjustbox}{max width=\textwidth,center}
\begin{tabular}{lcccc}
\toprule
\texttt{min\_move} & Aligned & Train/Val/Test & Used & Embargo \\
\midrule
0     & 4839 & 3420/630/699 & 4749 & 1.86\% \\
0.002 & 3666 & 2580/480/516 & 3576 & 2.45\% \\
0.004 & 2696 & 1890/330/386 & 2606 & 3.34\% \\
0.006 & 1989 & 1380/240/279 & 1899 & 4.52\% \\
0.008 & 1471 & 1050/150/181 & 1381 & 6.12\% \\
0.010 & 1054 & 750/90/124  & 964  & 8.54\% \\
\bottomrule
\end{tabular}
\end{adjustbox}
\end{table}

\section{Training Protocol and Reproducibility Details}
\label{app:train_proto}

All trainable baselines are implemented in PyTorch and trained under a unified optimization and checkpoint-selection protocol unless stated otherwise.

\paragraph{Optimization and checkpoint selection}
We use AdamW and select the best checkpoint by the chosen validation objective (default: MCC computed from thresholded probabilities at 0.5; \ref{app:eval_details}).
We report mean$\pm$std over multiple random seeds as specified in \ref{app:eval_details}.

\paragraph{Data loading and leakage-avoiding normalization}
All images are resized to $256\times256$ and loaded in raw $[0,1]$ space.
When enabled (\texttt{use\_norm}), we apply per-channel standardization using $(\boldsymbol{\mu},\boldsymbol{\sigma})$ estimated on the training split only; see \ref{app:input_param}.
Two-view inputs follow the fixed view-to-channel mapping in \Cref{sec:view-to-channel-mapping}.

\paragraph{Early stopping}
We train for at most 40 epochs with early stopping (\texttt{patience}=8).
We use a model-dependent minimum training epoch to stabilize optimization: \texttt{min\_epoch}=15 for \texttt{Lite-CNN} and \texttt{Lite-CNN-cons}, and \texttt{min\_epoch}=2 for \texttt{ResNet18-P} and \texttt{ResNet18-P-cons}.

\paragraph{Adversarial evaluation}
Adversarial evaluation is performed at test time only under the view-aligned protocol (\Cref{sec:protocol_main}).
Attack hyperparameters and budget conversion for standardized inputs are provided in \ref{app:attack_params} and \ref{app:input_param}.

\paragraph{Reproducibility}
We fix random seeds for Python, NumPy, and PyTorch and log the full configuration of each run, including the selected checkpoint.

\section{Attack Hyperparameters and Implementation Notes}
\label{app:attack_params}

\paragraph{Perturbation budgets}
Unless stated otherwise, perturbation budgets are defined in raw $[0,1]$ pixel space under an $\ell_\infty$ constraint, following standard adversarial-example evaluations~\citep{szegedy2014intriguing,goodfellow2014explaining,madry2017towards}.
We evaluate $\epsilon_{\mathrm{adv}}\in E$ as defined in Eq.~\eqref{eq:eps_grid_main}, where $\epsilon_{\mathrm{adv}}=0$ corresponds to clean evaluation.
When sweeping budgets, we run the evaluation once per $\epsilon_{\mathrm{adv}}$.
\Cref{tab:attack_params} summarizes the attack settings.

\paragraph{Loss and attack objective}
All attacks are \emph{untargeted} and maximize the cross-entropy loss with respect to the true label:
\[
\max_{\|\boldsymbol{\delta}\|_\infty\le \epsilon_{\mathrm{adv}}}\ \mathcal{L}_{\mathrm{CE}}\!\left(f(\mathbf{x}+\boldsymbol{\delta}),y\right).
\]
Here $\mathbf{x}$ denotes the raw $[0,1]$ input; when the model consumes standardized inputs, we optimize in the standardized space using the converted budget in Eq.~\ref{eq:app_eps_convert}.

\paragraph{FGSM (one-step)}
FGSM is implemented as a single-step update followed by clipping to the valid input range~\citep{goodfellow2014explaining}.

\paragraph{PGD (multi-step)}
PGD performs $K=10$ iterative updates with step size $\alpha=\epsilon_{\mathrm{adv}}/K$; we use a deterministic initialization (random\_start$=$False) from the clean input, and project after each step to enforce the $\ell_\infty$ constraint~\citep{madry2017towards}.

\begin{table}[!htbp]
\caption{Attack settings used in our experiments. Budgets are defined in raw $[0,1]$ space. When attacking standardized inputs, we convert $(\epsilon_{\mathrm{adv}},\alpha)$ using Eq.~\ref{eq:app_eps_convert}.}
\centering
\label{tab:attack_params}
\small
\begin{adjustbox}{max width=\textwidth,center}
\begin{tabular}{@{}llllll@{}}
\toprule
\textbf{Attack} & \textbf{Objective} & \textbf{Budget} & \textbf{Steps $K$} & \textbf{Step size $\alpha$} & \textbf{Random init.} \\
\midrule
FGSM & untargeted CE (maximize) & $\epsilon_{\mathrm{adv}}\in E$ & 1 & $\alpha=\epsilon_{\mathrm{adv}}$ & No \\
PGD  & untargeted CE (maximize) & $\epsilon_{\mathrm{adv}}\in E$ & 10 & $\alpha=\epsilon_{\mathrm{adv}}/10$ & No \\
\bottomrule
\end{tabular}
\end{adjustbox}
\end{table}

\section{Clean Baselines: Full Numerical Commentary}
\label{app:clean_baselines_full}

This appendix provides the full numerical commentary for \Cref{subsec:baseline} (clean evaluation), without changing any reported values or conclusions.
Representative values are summarized in \Cref{tab:baseline_mcc_key}; full curves are shown in \Cref{fig:baseline_minmove}.

\paragraph{Minimum-movement filtering defines the evaluation regime}
\Cref{fig:baseline_minmove} shows that the minimum-movement threshold $\tau$ (\Cref{subsec:minmove}) largely determines whether the task is learnable \emph{on the ex-post filtered subsets} defined by \texttt{min\_move}.
Without filtering or with weak filtering ($\tau\in\{0,0.002,0.004\}$), most deep models hover around zero MCC (sometimes negative), indicating near-chance behavior on the full set.
As $\tau$ increases to $0.006$ and $0.008$, performance improves sharply, consistent with restricting evaluation to subsets that exclude near-zero moves and thus reduce direction-label ambiguity.
Aggregating across all deep configurations (Lite-CNN/ResNet18-P and \texttt{*-cons}, all input modes), the mean MCC rises from $-0.002$ at $\tau=0$ to $0.082$ at $\tau=0.006$ and peaks at $0.092$ at $\tau=0.008$, before dropping to $0.030$ at $\tau=0.010$.
The drop at $\tau=0.010$ aligns with the reduced effective sample size (\Cref{tab:minmove_embargo_effect}), which increases variance and destabilizes training.
We stress that $\tau$ is used here to define evaluation regimes ex post, not as an inference-time filtering rule.

\paragraph{Lite-CNN: single-view is more robust under weak filtering; dual-view becomes beneficial once labels stabilize}
Under weak filtering ($\tau=0.004$), \texttt{indic} already yields a clear positive signal for Lite-CNN ($0.107292\pm0.022477$), while \texttt{both} early fusion is negative ($-0.020004\pm0.016396$), suggesting negative transfer when two noisy views are fused too early.
Once $\tau\ge 0.006$, \texttt{both} becomes consistently positive and competitive (e.g., $0.088767\pm0.042662$ at $\tau=0.006$, $0.107206\pm0.046632$ at $\tau=0.008$), indicating that dual-view information is exploitable when labels are less noisy.
At $\tau=0.006$, \texttt{ohlcv} is also strong ($0.109095\pm0.039023$), implying that price-based visual patterns alone can be predictive after filtering.

\paragraph{Late fusion drives the main gain; consistency regularization yields modest, non-monotonic effects}
Switching from early fusion (\texttt{lite-cnn | both}) to dual-branch late fusion without consistency ($\lambda=0$) yields a large improvement at $\tau=0.006$ (from $0.088767\pm0.042662$ to $0.127459\pm0.035946$), indicating that the \emph{architecture-level} change (separate encoders + fusion head) is the dominant factor in our setting.
Adding consistency regularization can help, but not monotonically and the gains are modest: at $\tau=0.006$, \texttt{lite-cnn-cons} peaks at $\lambda=0.5$ ($0.129607\pm0.026449$), while $\lambda=1$ is lower ($0.120755\pm0.029468$).
At $\tau=0.008$, $\lambda=1$ reaches $0.103269\pm0.031387$, close to early fusion ($0.107206\pm0.046632$), suggesting that consistency appears to act largely as a regularizer, whose benefit is \emph{$\tau$-dependent} and sensitive to the effective data regime induced by filtering.

\paragraph{ResNet18-P: early fusion underperforms; late fusion partially recovers but variance is higher}
ResNet18-P exhibits a different regime: dual-view early fusion is consistently weak (e.g., $0.042873\pm0.061370$ at $\tau=0.008$), while single-view modes are stronger at the same threshold (\texttt{indic}: $0.108023\pm0.040145$, \texttt{ohlcv}: $0.109101\pm0.072830$).
Late fusion improves the dual-view setting (e.g., \texttt{resnet18p-cons} with $\lambda=0.5$ reaches $0.083423\pm0.052584$ at $\tau=0.008$), but the ResNet family generally shows larger standard deviations, especially under strict filtering (e.g., \texttt{resnet18p | both}: $0.061769\pm0.115898$ at $\tau=0.010$), consistent with sample-size-driven instability.

\paragraph{LogReg as a diagnostic baseline}
LogReg can be competitive at certain thresholds (e.g., \texttt{ohlcv} at $\tau=0.008$ achieves $\mathrm{MCC}=0.116126$), but its behavior is highly non-monotonic across $\tau$ and input modes (e.g., \texttt{both} at $\tau=0.006$ yields $\mathrm{MCC}=-0.039129$).
We therefore treat it primarily as a diagnostic indicator of partial linear separability after label stabilization.

% ==================== Restored full branch diagnostics appendix ====================
\section{Full Branch Diagnostics for Late-Fusion Models}
\label{app:branch_diag_full}

\paragraph{Branch diagnostics for late-fusion two-view models (clean test, $\tau=0.006$)}
To interpret robustness trends for \texttt{*-cons} models, we inspect clean-test branch diagnostics that summarize
cross-branch alignment (\texttt{agree\_ab}, \texttt{sym\_kl} with $T{=}2$) and branch-wise versus fused MCC
(\Cref{tab:branch_diag_compact}; full results in Tables~\ref{tab:branch_diag_full_conf}--\ref{tab:branch_mcc_full}).
For \texttt{Lite-CNN-Cons}, enabling consistency ($\lambda>0$) sharply increases alignment (higher \texttt{agree\_ab}, much lower \texttt{sym\_kl}),
yet branch-wise MCC remains near zero while fused-head MCC stays substantially higher.
This indicates that predictive signal is expressed primarily through the \emph{feature-level late-fusion pathway} rather than strong standalone branch classifiers.
For \texttt{ResNet18P-Cons}, larger $\lambda$ likewise improves alignment and is associated with higher fused MCC, but variance remains notable and robustness failures under indicator-view attacks persist.

\begin{table}[!htbp]
  \caption{\textbf{Compact clean-test diagnostics for late-fusion two-view models ($\tau=0.006$, $T=2$).}
  Mean$\pm$std over $n{=}8$ seeds. We report cross-branch alignment (\texttt{agree\_ab}, \texttt{sym\_kl}),
  branch-wise MCC (\texttt{mcc\_a}/\texttt{mcc\_b}), and fused-head MCC (\texttt{mcc\_fuse}). Full diagnostics are in Tables~\ref{tab:branch_diag_full_conf}--\ref{tab:branch_mcc_full}}
  \label{tab:branch_diag_compact}
  \centering
  \scriptsize
  \setlength{\tabcolsep}{4pt}
  \renewcommand{\arraystretch}{1.05}
  \begin{adjustbox}{max width=\textwidth,center}
\begin{tabular}{lcrrrr}
    \toprule
    \textbf{Model} & $\boldsymbol{\lambda}$ & \textbf{agree\_ab} & \textbf{sym\_kl} & \textbf{mcc\_a / mcc\_b} & \textbf{mcc\_fuse} \\
    \midrule
    Lite-CNN-Cons  & 0   & $0.275090\pm0.244057$ & $0.164898\pm0.125751$ &
    $0.019377\pm0.061519\ /\ 0.005906\pm0.051856$ & $0.127459\pm0.035946$ \\
    Lite-CNN-Cons  & 0.5 & $0.830197\pm0.185333$ & $0.000195\pm0.000098$ &
    $0.001488\pm0.053262\ /\ -0.001753\pm0.076580$ & $0.129607\pm0.026449$ \\
    Lite-CNN-Cons  & 1   & $0.803315\pm0.205572$ & $0.000106\pm0.000029$ &
    $0.006285\pm0.057660\ /\ -0.015335\pm0.052686$ & $0.120755\pm0.029468$ \\
    \midrule
    ResNet18P-Cons & 0   & $0.439964\pm0.247314$ & $0.074461\pm0.065610$ &
    $0.014477\pm0.047350\ /\ 0.000779\pm0.038367$ & $0.048051\pm0.056600$ \\
    ResNet18P-Cons & 0.5 & $0.920251\pm0.196921$ & $0.006523\pm0.003518$ &
    $-0.006042\pm0.012812\ /\ -0.002165\pm0.015076$ & $0.067112\pm0.031388$ \\
    ResNet18P-Cons & 1   & $0.931452\pm0.163280$ & $0.003425\pm0.001379$ &
    $-0.002706\pm0.028186\ /\ 0.009707\pm0.019310$ & $0.095146\pm0.057933$ \\
    \bottomrule
  \end{tabular}
\end{adjustbox}
\end{table}

Tables~\ref{tab:branch_diag_full_conf}--\ref{tab:branch_mcc_full} provide confidence, class-probability means, and branch-vs-fused MCC, respectively.

% -------------------- (B) Confidence diagnostics (conf_*) --------------------
\begin{table}[!p]
  \caption{\textbf{Full clean-test confidence diagnostics for late-fusion two-view models ($\tau=0.006$, $T=2$).}
  Mean$\pm$std over $n{=}8$ seeds. \texttt{conf\_*} denotes mean max-probability (branch confidences computed on temperature-scaled distributions with $T=2$; fused confidence on $\mathrm{softmax}(z_f)$).}
  \label{tab:branch_diag_full_conf}
  \centering
  \small
  \setlength{\tabcolsep}{6pt}
  \renewcommand{\arraystretch}{1.05}
  \begin{adjustbox}{max width=\textwidth,center}
\begin{tabular}{lcrrr}
    \toprule
    \textbf{Model} & $\boldsymbol{\lambda}$ & \textbf{conf\_a} & \textbf{conf\_b} & \textbf{conf\_f} \\
    \midrule
    Lite-CNN-Cons  & 0   & $0.627352\pm0.070257$ & $0.554910\pm0.020764$ & $0.917221\pm0.007478$ \\
    Lite-CNN-Cons  & 0.5 & $0.508533\pm0.004537$ & $0.506999\pm0.003842$ & $0.913617\pm0.005166$ \\
    Lite-CNN-Cons  & 1   & $0.507229\pm0.004763$ & $0.507287\pm0.005493$ & $0.913873\pm0.007414$ \\
    \midrule
    ResNet18P-Cons & 0   & $0.570899\pm0.043481$ & $0.567578\pm0.040866$ & $0.782159\pm0.078004$ \\
    ResNet18P-Cons & 0.5 & $0.605562\pm0.065867$ & $0.611589\pm0.066048$ & $0.802972\pm0.056561$ \\
    ResNet18P-Cons & 1   & $0.622989\pm0.083976$ & $0.624896\pm0.082731$ & $0.808066\pm0.048449$ \\
    \bottomrule
  \end{tabular}
\end{adjustbox}
\end{table}

% -------------------- (C) Positive-class mean diagnostics (pos_mean_*) --------------------
\begin{table}[!p]
  \caption{\textbf{Full clean-test positive-class probability means for late-fusion two-view models ($\tau=0.006$, $T=2$).}
  Mean$\pm$std over $n{=}8$ seeds. \texttt{pos\_mean\_*} denotes the mean positive-class probability (branch means computed on temperature-scaled distributions with $T=2$; fused mean on $\mathrm{softmax}(z_f)$).}
  \label{tab:branch_diag_full_pos}
  \centering
  \small
  \setlength{\tabcolsep}{6pt}
  \renewcommand{\arraystretch}{1.05}
  \begin{adjustbox}{max width=\textwidth,center}
\begin{tabular}{lcrrr}
    \toprule
    \textbf{Model} & $\boldsymbol{\lambda}$ & \textbf{pos\_mean\_a} & \textbf{pos\_mean\_b} & \textbf{pos\_mean\_f} \\
    \midrule
    Lite-CNN-Cons  & 0   & $0.522772\pm0.145785$ & $0.511727\pm0.058662$ & $0.514203\pm0.039776$ \\
    Lite-CNN-Cons  & 0.5 & $0.502299\pm0.009459$ & $0.499700\pm0.008371$ & $0.511911\pm0.056243$ \\
    Lite-CNN-Cons  & 1   & $0.498687\pm0.008587$ & $0.498635\pm0.009367$ & $0.520245\pm0.068696$ \\
    \midrule
    ResNet18P-Cons & 0   & $0.549286\pm0.065936$ & $0.513286\pm0.074454$ & $0.455765\pm0.133288$ \\
    ResNet18P-Cons & 0.5 & $0.592447\pm0.085277$ & $0.590547\pm0.095968$ & $0.483531\pm0.135293$ \\
    ResNet18P-Cons & 1   & $0.564726\pm0.139656$ & $0.564738\pm0.140885$ & $0.482701\pm0.169748$ \\
    \bottomrule
  \end{tabular}
\end{adjustbox}
\end{table}

% -------------------- (D) Branch-wise vs fused MCC --------------------
\begin{table}[!p]
  \caption{\textbf{Branch-wise vs.\ fused MCC on clean test ($\tau=0.006$, threshold $0.5$).}
  Mean$\pm$std over $n{=}8$ seeds.}
  \label{tab:branch_mcc_full}
  \centering
  \small
  \setlength{\tabcolsep}{6pt}
  \renewcommand{\arraystretch}{1.05}
  \begin{adjustbox}{max width=\textwidth,center}
\begin{tabular}{lcrrr}
    \toprule
    \textbf{Model} & $\boldsymbol{\lambda}$ & \textbf{mcc\_a} & \textbf{mcc\_b} & \textbf{mcc\_fuse} \\
    \midrule
    Lite-CNN-Cons  & 0   & $0.019377\pm0.061519$ & $0.005906\pm0.051856$  & $0.127459\pm0.035946$ \\
    Lite-CNN-Cons  & 0.5 & $0.001488\pm0.053262$ & $-0.001753\pm0.076580$ & $0.129607\pm0.026449$ \\
    Lite-CNN-Cons  & 1   & $0.006285\pm0.057660$ & $-0.015335\pm0.052686$ & $0.120755\pm0.029468$ \\
    \midrule
    ResNet18P-Cons & 0   & $0.014477\pm0.047350$ & $0.000779\pm0.038367$  & $0.048051\pm0.056600$ \\
    ResNet18P-Cons & 0.5 & $-0.006042\pm0.012812$ & $-0.002165\pm0.015076$ & $0.067112\pm0.031388$ \\
    ResNet18P-Cons & 1   & $-0.002706\pm0.028186$ & $0.009707\pm0.019310$  & $0.095146\pm0.057933$ \\
    \bottomrule
  \end{tabular}
\end{adjustbox}
\end{table}

\section{Sensitivity to the Operating Point: Adversarial Robustness at $\tau=0.008$}
\label{app:tau008}

This appendix reports adversarial robustness results under a larger min\_move threshold $\tau=0.008$.
Because changing $\tau$ alters the labeling rule (and typically class balance), absolute clean MCC values should not be compared across $\tau$; we therefore focus on within-$\tau$ degradation patterns.

\begin{table}[!htbp]
  \caption{\textbf{PGD robustness summary at $\tau=0.008$ (min\_move).}
  We report MCC at two representative budgets ($0.25/255$ and $1/255$), the worst-case drop $\Delta_{\mathrm{worst}}$ over nonzero budgets $\{0.25,0.5,0.75,1\}/255$ (Eq.~\eqref{eq:worst_main}),
  and the stress-test MCC at $\epsilon_{\mathrm{adv}}=2/255$.}
  \label{tab:adv_summary_pgd_tau008}
  \centering
  \scriptsize
  \setlength{\tabcolsep}{4pt}
  \begin{adjustbox}{max width=\textwidth,center}
\begin{tabular}{lllrrrrr}
    \toprule
    \textbf{Model} & \textbf{Input} & \textbf{Attack (PGD)} &
    \textbf{MCC@0} & \textbf{MCC@0.25/255} & \textbf{MCC@1/255} & $\boldsymbol{\Delta_{\mathrm{worst}}}$ & \textbf{MCC@2/255} \\
    \midrule
    Lite-CNN & \texttt{ohlcv} & on \texttt{ohlcv} & 0.091 & 0.042 & -0.101 & -0.192 & -0.279 \\
    Lite-CNN & \texttt{indic} & on \texttt{indic} & 0.100 & 0.035 & -0.187 & -0.287 & -0.410 \\
    \midrule
    Lite-CNN & \texttt{both} (early) & on \texttt{ohlcv} (ch0) & 0.107 & 0.047 & -0.101 & -0.209 & -0.300 \\
    Lite-CNN & \texttt{both} (early) & on \texttt{indic} (ch1) & 0.107 & 0.043 & -0.101 & -0.208 & -0.272 \\
    Lite-CNN & \texttt{both} (early) & on both & 0.107 & -0.008 & -0.285 & -0.393 & -0.561 \\
    \midrule
    Lite-CNN-Cons ($\lambda{=}0$) & \texttt{both} (late) & on \texttt{ohlcv} (ch0) & 0.100 & 0.061 & -0.048 & -0.148 & -0.198 \\
    Lite-CNN-Cons ($\lambda{=}0$) & \texttt{both} (late) & on \texttt{indic} (ch1) & 0.100 & 0.087 & 0.058 & -0.043 & 0.032 \\
    Lite-CNN-Cons ($\lambda{=}0$) & \texttt{both} (late) & on both & 0.100 & 0.057 & -0.078 & -0.178 & -0.249 \\
    \midrule
    Lite-CNN-Cons ($\lambda{=}0.5$) & \texttt{both} (late) & on \texttt{ohlcv} (ch0) & 0.090 & 0.062 & -0.051 & -0.141 & -0.179 \\
    Lite-CNN-Cons ($\lambda{=}0.5$) & \texttt{both} (late) & on \texttt{indic} (ch1) & 0.090 & 0.084 & 0.063 & -0.027 & 0.043 \\
    Lite-CNN-Cons ($\lambda{=}0.5$) & \texttt{both} (late) & on both & 0.090 & 0.053 & -0.080 & -0.170 & -0.246 \\
    \midrule
    Lite-CNN-Cons ($\lambda{=}1$) & \texttt{both} (late) & on \texttt{ohlcv} (ch0) & 0.103 & 0.059 & -0.070 & -0.173 & -0.219 \\
    Lite-CNN-Cons ($\lambda{=}1$) & \texttt{both} (late) & on \texttt{indic} (ch1) & 0.103 & 0.099 & 0.064 & -0.039 & 0.018 \\
    Lite-CNN-Cons ($\lambda{=}1$) & \texttt{both} (late) & on both & 0.103 & 0.042 & -0.098 & -0.201 & -0.269 \\
    \midrule
    ResNet18-P & \texttt{ohlcv} & on \texttt{ohlcv} & 0.109 & -0.489 & -0.977 & -1.087 & -0.999 \\
    ResNet18-P & \texttt{indic} & on \texttt{indic} & 0.108 & -1.000 & -1.000 & -1.108 & -1.000 \\
    \midrule
    ResNet18-P & \texttt{both} (early) & on \texttt{ohlcv} (ch0) & 0.043 & -0.453 & -0.962 & -1.005 & -1.000 \\
    ResNet18-P & \texttt{both} (early) & on \texttt{indic} (ch1) & 0.043 & -0.830 & -1.000 & -1.043 & -1.000 \\
    ResNet18-P & \texttt{both} (early) & on both & 0.043 & -0.936 & -1.000 & -1.043 & -1.000 \\
    \midrule
    ResNet18-P-Cons ($\lambda{=}0$) & \texttt{both} (late) & on \texttt{ohlcv} (ch0) & 0.079 & -0.240 & -0.799 & -0.878 & -0.974 \\
    ResNet18-P-Cons ($\lambda{=}0$) & \texttt{both} (late) & on \texttt{indic} (ch1) & 0.079 & -0.927 & -1.000 & -1.079 & -1.000 \\
    ResNet18-P-Cons ($\lambda{=}0$) & \texttt{both} (late) & on both & 0.079 & -0.966 & -1.000 & -1.079 & -1.000 \\
    \midrule
    ResNet18-P-Cons ($\lambda{=}0.5$) & \texttt{both} (late) & on \texttt{ohlcv} (ch0) & 0.083 & -0.246 & -0.791 & -0.875 & -0.986 \\
    ResNet18-P-Cons ($\lambda{=}0.5$) & \texttt{both} (late) & on \texttt{indic} (ch1) & 0.083 & -0.921 & -1.000 & -1.083 & -1.000 \\
    ResNet18-P-Cons ($\lambda{=}0.5$) & \texttt{both} (late) & on both & 0.083 & -0.964 & -1.000 & -1.083 & -1.000 \\
    \midrule
    ResNet18-P-Cons ($\lambda{=}1$) & \texttt{both} (late) & on \texttt{ohlcv} (ch0) & 0.094 & -0.247 & -0.795 & -0.889 & -0.974 \\
    ResNet18-P-Cons ($\lambda{=}1$) & \texttt{both} (late) & on \texttt{indic} (ch1) & 0.094 & -0.944 & -1.000 & -1.094 & -1.000 \\
    ResNet18-P-Cons ($\lambda{=}1$) & \texttt{both} (late) & on both & 0.094 & -0.978 & -1.000 & -1.094 & -1.000 \\
    \bottomrule
\end{tabular}
\end{adjustbox}
\end{table}

\begin{table}[!htbp]
  \caption{\textbf{Compact clean-test diagnostics for late-fusion two-view models at $\tau=0.008$ ($T=2$).}
  Mean$\pm$std over $n{=}8$ seeds. We report cross-branch alignment (\texttt{agree\_ab}, \texttt{sym\_kl}),
  branch-wise MCC (\texttt{mcc\_a}/\texttt{mcc\_b}), and fused-head MCC (\texttt{mcc\_fuse}).}
  \label{tab:branch_diag_compact_tau008}
  \centering
  \scriptsize
  \setlength{\tabcolsep}{4pt}
  \renewcommand{\arraystretch}{1.05}
  \begin{adjustbox}{max width=\textwidth,center}
\begin{tabular}{lcrrrr}
    \toprule
    \textbf{Model} & $\boldsymbol{\lambda}$ & \textbf{agree\_ab} & \textbf{sym\_kl} & \textbf{mcc\_a / mcc\_b} & \textbf{mcc\_fuse} \\
    \midrule
    Lite-CNN-Cons & 0 & $0.470994\pm0.357293$ & $0.119371\pm0.092575$ & $0.005352\pm0.102980$\ /\ $-0.017188\pm0.039365$ & $0.100430\pm0.034439$ \\
    Lite-CNN-Cons & 0.5 & $0.775552\pm0.170337$ & $0.000164\pm0.000088$ & $-0.009824\pm0.052837$\ /\ $-0.004969\pm0.047469$ & $0.089973\pm0.067209$ \\
    Lite-CNN-Cons & 1 & $0.778315\pm0.208975$ & $0.000191\pm0.000099$ & $-0.003662\pm0.048320$\ /\ $-0.007522\pm0.051147$ & $0.103269\pm0.031387$ \\
    ResNet18P-Cons & 0 & $0.408840\pm0.195311$ & $0.061022\pm0.040954$ & $0.014461\pm0.048496$\ /\ $0.024219\pm0.084118$ & $0.078607\pm0.054662$ \\
    ResNet18P-Cons & 0.5 & $0.946823\pm0.150407$ & $0.003367\pm0.001974$ & $-0.027162\pm0.076826$\ /\ $-0.008803\pm0.024899$ & $0.083423\pm0.052584$ \\
    ResNet18P-Cons & 1 & $0.934392\pm0.112509$ & $0.005748\pm0.003453$ & $-0.016075\pm0.030005$\ /\ $-0.043617\pm0.080173$ & $0.093970\pm0.101626$ \\
    \bottomrule
\end{tabular}
\end{adjustbox}
\end{table}

\section{PGD Stress Test (up to \texorpdfstring{$\epsilon_{\mathrm{adv}}=2/255$}{epsilon\_adv=2/255})}
\label{app:stress_2over255}

To complement the main robustness grid (which uses $\epsilon_{\mathrm{adv}}\le 1/255$), we extend the perturbation budget to $\epsilon_{\mathrm{adv}}=2/255$ for the late-fusion two-view Lite-CNN-Cons models.
This stress test is evaluated at both $\tau=0.006$ and $\tau=0.008$.
Across both operating points, Lite-CNN-Cons remains \emph{non-trivially robust} under the view-constrained \texttt{indic}-only attack (ch1) even at $2/255$ (best mean MCC $\approx 0.048\pm0.059$ at $\tau=0.006$ with $\lambda=1$, and $\approx 0.043\pm0.047$ at $\tau=0.008$ with $\lambda=0.5$).
In contrast, attacking \texttt{ohlcv} only (ch0) already drives MCC negative at the same budget (best mean MCC $\approx -0.199\pm0.100$ at $\tau=0.006$ and $\approx -0.179\pm0.117$ at $\tau=0.008$), and the joint two-view attack remains the hardest case (best mean MCC $\approx -0.278\pm0.092$ at $\tau=0.006$ and $\approx -0.246\pm0.078$ at $\tau=0.008$).
These results strengthen the interpretation that late fusion primarily helps when the attacker cannot manipulate both views (redundancy across branches), while it cannot guarantee robustness under joint perturbations.

\paragraph{Random-start check}
We also verified that enabling \texttt{random\_start} (with multiple restarts) does not materially change the results at $2/255$ (we did not observe worse MCC), which is consistent with a locally near-linear loss landscape where PGD quickly converges to the same boundary (sign-gradient) solution regardless of initialization.

\begin{table}[!htbp]
  \caption{\textbf{PGD stress test at $\epsilon_{\mathrm{adv}}=2/255$.}
  We report MCC on clean data and under three PGD threat models restricted to \texttt{indic} (ch1), \texttt{ohlcv} (ch0), or both views.}
  \label{tab:pgd_stress_2over255}
  \centering
  \small
  \setlength{\tabcolsep}{6pt}
  \begin{adjustbox}{max width=\textwidth,center}
  \begin{tabular}{ccrrrr}
    \toprule
    $\boldsymbol{\tau}$ & $\boldsymbol{\lambda}$ & \textbf{MCC@0} &
    \textbf{MCC@2/255 (ch1)} & \textbf{MCC@2/255 (ch0)} & \textbf{MCC@2/255 (both)} \\
    \midrule
0.006 & 0 & 0.127459±0.035946 & 0.029360±0.045407 & -0.202645±0.076818 & -0.285711±0.069225 \\
0.006 & 0.5 & 0.129607±0.026449 & 0.030246±0.067472 & -0.215792±0.097168 & -0.310946±0.094411 \\
0.006 & 1 & 0.120755±0.029468 & 0.048053±0.058530 & -0.198704±0.099514 & -0.277539±0.092451 \\
\midrule
0.008 & 0 & 0.100430±0.034439 & 0.031542±0.037874 & -0.198313±0.070048 & -0.248958±0.082638 \\
0.008 & 0.5 & 0.089973±0.067209 & 0.042798±0.046734 & -0.178680±0.116959 & -0.246481±0.078135 \\
0.008 & 1 & 0.103269±0.031387 & 0.018316±0.048887 & -0.218880±0.104764 & -0.268799±0.074658 \\
    \bottomrule
  \end{tabular}
  \end{adjustbox}
\end{table}

\end{document}